\documentclass[sigconf]{acmart}
\usepackage{url}
\usepackage[utf8]{inputenc}
\usepackage{subcaption}
\usepackage{graphicx}
\usepackage{amsmath}
\usepackage{booktabs}
\usepackage[utf8]{inputenc}

\DeclareUnicodeCharacter{1EF3}{\`y}
\DeclareUnicodeCharacter{2212}{-}
\DeclareUnicodeCharacter{03C0}{$\pi$}

\usepackage{cleveref}
\usepackage{hyperref}
\usepackage{comment}

\usepackage{todonotes}
\usepackage{natbib}
\usepackage{adjustbox}
\usepackage{fancyhdr}
\usepackage[frozencache=true, cachedir=minted-cache]{minted}
\usepackage{spverbatim}

\AtBeginDocument{%
  }

\copyrightyear{2022}
\acmYear{2022}
\setcopyright{rightsretained}
\acmConference[CIKM '22]{Proceedings of the 31st ACM International Conference on Information and Knowledge Management}{October 17--21, 2022}{Atlanta, GA, USA}
\acmBooktitle{Proceedings of the 31st ACM International Conference on Information and Knowledge Management (CIKM '22), October 17--21, 2022, Atlanta, GA, USA}
\acmDOI{10.1145/3511808.3557079}
\acmISBN{978-1-4503-9236-5/22/10}

\copyrightyear{2022}
\acmYear{2022}
\setcopyright{rightsretained}
\acmConference[CIKM '22]{Proceedings of the 31st ACM International
Conference on Information and Knowledge Management}{October
17--21, 2022}{Atlanta, GA, USA}
\acmBooktitle{Proceedings of the 31st ACM International Conference on
Information and Knowledge Management (CIKM '22), October 17--21,
2022, Atlanta, GA, USA}
\acmDOI{10.1145/3511808.3557079}
\acmISBN{978-1-4503-9236-5/22/10}

\begin{document}

\title{Fooling MOSS Detection with Pretrained Language Models}

\author{Stella Biderman}
\email{biderman_stella@bah.com}
\affiliation{%
  \institution{Booz Allen Hamilton}
  \institution{EleutherAI}
  \country{USA}
}

\author{Edward Raff}
\email{raff_edward@bah.com}
\affiliation{%
  \institution{Booz Allen Hamilton}
  \country{USA}
}

\begin{abstract}
  As artificial intelligence (AI) technologies become increasingly powerful and prominent in society, their misuse is a growing concern. In educational settings, AI technologies could be used by students to cheat on assignments and exams. In this paper we explore whether transformers can be used to solve introductory level programming assignments while bypassing commonly used AI tools to detect similarities between pieces of software. We find that a student using GPT-J \citep{gpt-j} can complete introductory level programming assignments without triggering suspicion from MOSS \citep{aiken2000moss}, a widely used software similarity and plagiarism detection tool. This holds despite the fact that GPT-J was not trained on the problems in question and is not provided with any examples to work from. We further find that the code written by GPT-J is diverse in structure, lacking any particular tells that future plagiarism detection techniques may use to try to identify algorithmically generated code. We conclude with a discussion of the ethical and educational implications of large language models and directions for future research.
\end{abstract}



\maketitle

\section{Introduction}
The COVID-19 pandemic has lead to a boom in online education \citep{emergency} due to the health risks of in-person teaching \citep{juliana2020turnitin}. This has in turn lead to a growing market for tools that claim to ensure academic integrity in online courses \citep{yasid2020plagiarism}. With the increased attention to and wider use of these tools, it becomes even more important to have thorough understanding of if, and how, such tools may be subverted by students. While there is a substantial literature on plagiarism techniques and detection strategies, this literature has not yet engaged with the recent breakthroughs in transformers research. Similarly, recent work in the transformers literature has investigated the ability of large language models to solve college-level homework assignments in a variety of domains including calculus \citep{drori2021neural}, linear algebra \citep{drori2021solving}, astronomy \citep{shporerlearning}, machine learning \citep{tran2021solving} but has not engaged with the issues and implications of real-world application of their results.

\begin{figure}[ht]
\centering
\adjustbox{max width=\columnwidth}{
\includegraphics{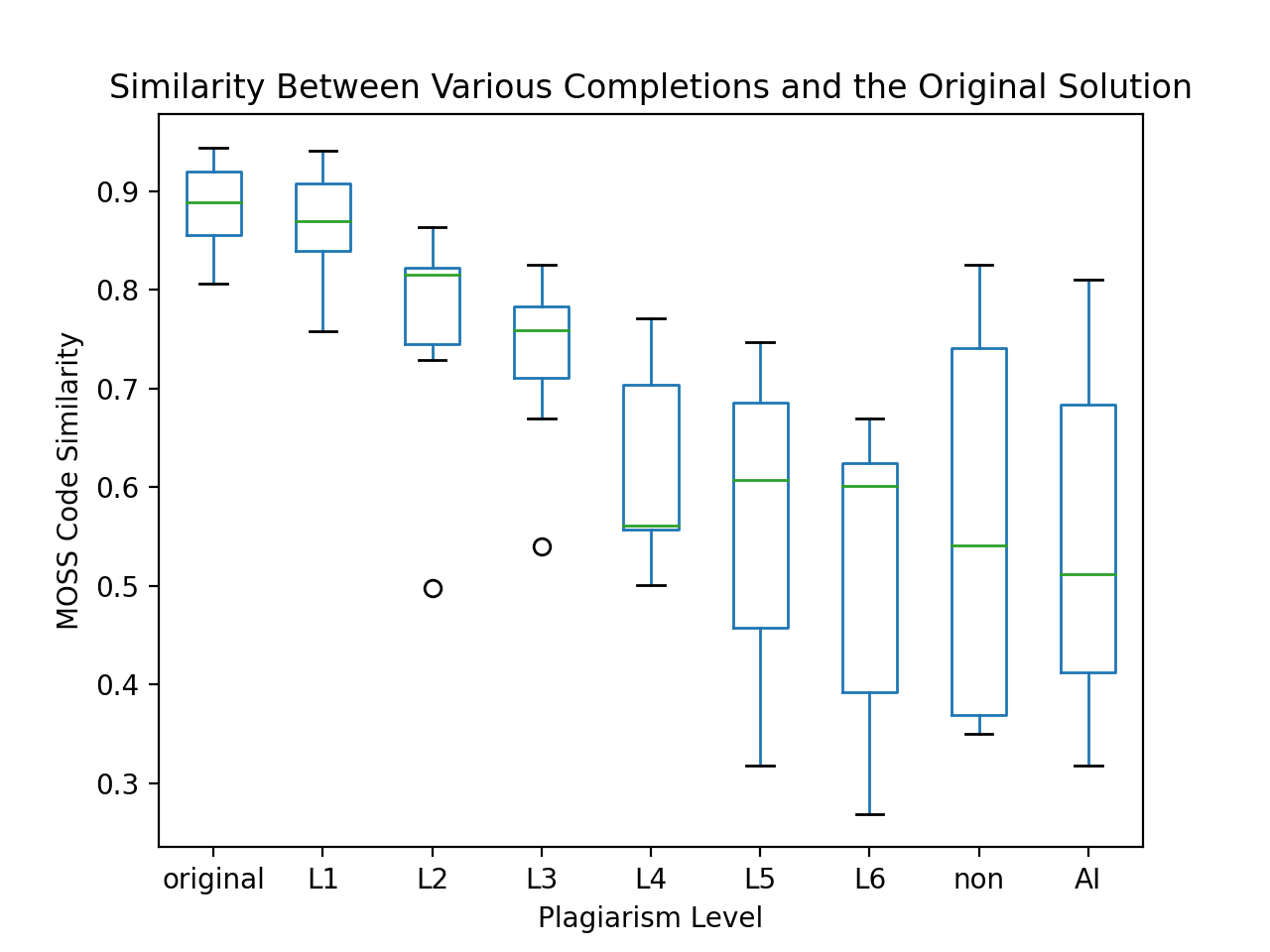}
}
\caption{A breakdown of MOSS scores for completions of exercises from \citet{karnalim2019source} by how they were generated.  ``Original'' denotes the original solutions, ``LN'' denotes Level N plagiarism, ``non'' denotes non-plagiarisms, and ``AI'' denotes solutions written by GPT-J.}\label{fig:moss_scores}
\end{figure}

In this paper we carry out the first investigation of the ability of a transformer-based language model to generate \textit{de novo} completions of homework assignments in a college-level course while bypassing plagiarism detection techniques. For our investigation we choose to use Measure of Software Similarity (MOSS) \citep{aiken2000moss}, a popular tool used to monitor academic integrity in universities worldwide \citep{bowyer1999experience,lancaster2004comparison,luke2014software,sheahen2016taps}, to identify likely plagiarisms. We choose to study MOSS because it is an open-source model representing the state-of-the-art for plagiarism detection and its design has been widely copied by commercial tools \citep{devore2020mossad}. While the licensing of commercial tools prevent us from studying them directly, we anticipate our findings to apply to the many commercial tools based on MOSS. We find that the GPT-J model \citep{gpt-j} is capable of fooling MOSS \citep{aiken2000moss} with a high degree of reliability, that little human editing is required to produce assignments that would receive full marks in a course, and that GPT-J solutions have sufficient diversity to avoid easy detection. Our study raises serious questions about the efficacy of using MOSS-based tools as a primary identifier of academic plagiarism due to the new risks posed by modern language models.

\subsection{Deployment Context}

Our study focuses on GPT-J and MOSS because they are both widely used and freely and publicly available to the public. GPT-J is the most powerful easily usable transformer model in the world, via the free API maintained by EleutherAI at \href{6b.eleuther.ai}{6b.eleuther.ai}. Although there exist more powerful transformers that are publicly available, they are either incapable of generating code \citep{raffel2019exploring,sanh2022multitask} or lack free public APIs and require substantial expertise to deploy \citep{black2022gpt,zhang2022opt}.

GPT-J and similar models have also been widely deployed in industry applications including for writing and code completion tasks. The API we use has been used over 3 million times and the family of autoregressive language models it comes from have been downloaded over 25 million times in total from the official hosting platforms on \href{https://the-eye.eu/public/AI/GPT-J-6B/}{the Eye} and \href{https://huggingface.co/EleutherAI}{HuggingFace}. The models have also been integrated into many AI products and services including ones by \href{https://kiel.ai/how-we-setup-gpt-neo-gpt-3-alternative/}{Kiel AI}, \href{https://www.pragnakalp.com/gpt-j-6b-parameters-model-huggingface/}{PragnaKalp}, \href{https://www.helm-nagel.com/open-sourced-gpt-Neo-writes-individualized-emails}{Helm \& Nagel GmbH}, \href{https://www.deepgenx.com/}{CodeGenX}, and \href{https://prototypr.io/home/}{Prototypr}. Beyond code generation, the models are also popular for generating emails\footnote{https://www.helm-nagel.com/open-sourced-gpt-Neo-writes-individualized-emails}, powering chatbots\footnote{https://writeholo.com/}, and creative writing\footnote{https://blog.coreweave.com/powering-a-world-of-possibility-with-gpt-j}.

MOSS has been widely used by professors across the world for decades, including at University of South Florida \citep{bowyer1999experience}, the Georgia Institute of Technology \citep{sheahen2016taps}, Bundeswehr University, Munich \citep{pawelczak2018benefits}, University of Newcastle \citep{karnalim2020choosing}, and Utah Valley University \citep{sheikhteaching}. In 2020 there were over 300K currently active MOSS accounts, growing at a rate of 50k-100k per year \citep{devore2020mossad}. MOSS is also the basis for or is integrated with a number of commercial tools for plagiarism detection, including \href{https://codequiry.com/moss/measure-of-software-similarity}{CodeQuiry} and GradeScope \citep{singh2017gradescope}.

\subsection{Outline}

The rest of our paper is organized as follows. First we will review the related work, and how prior language models have not provided satisfying answers to the plagiarism risk of language models in \autoref{sec:related_work}. Next we will review our methodology in \autoref{sec:methodology}, where we provide a realistic approach to how a student may use GPT-J to plagiarize an assignment, and what (simple) prompt engineering is required to produce the results. \autoref{sec:results} performs a detailed analysis of the results, cataloging the types of errors made by GPT-J, measures the similarity and detection risk of multiple plagiarism strategies, and shows how GPT-J is challenging to detect. Further detailed analysis of individual assignments is in the appendix. Due to the nature of this work, we discuss the ethics and why we believe the benefits outweigh their risks in \autoref{sec:ethics}, noting that the limits of GPT-J's abilities prevent it from allowing a student to proceed through an entire degree, and other teaching touch points like exams and quizzes provide a remediation for the informed instructor. Finally we conclude with a discussion and related Human-Computer Interaction (HCI) problems in \autoref{sec:conclusion}.

\section{Related Work} \label{sec:related_work}

Very recent concurrent works have looked at college assignment generation \citep{drori2021neural,shporerlearning,tran2021solving} as a means of understanding what language models like GPT-3 \citep{gpt3} and Codex \citep{chen2021evaluating} learn. Although these models are substantially more powerful than GPT-J, the study setups do not evaluate a realistic scenario for plagiarism. Significant coding, prompt engineering, and infrastructure is required for the experiments that a novice student would not have or be able to replicate (indeed if they could, there is no question they could complete these introductory assignments without assistance). Our work instead focus on a realistic threat model to determine that there is a real risk that needs further study, though we suspect many of the best solutions would be non-technical (See \autoref{sec:ethics}). Other recent work has saught to develop code models that are accessible to students \citep{fried2022incoder,xu2022systematic}, but these models are pure code models and therefore lack the ability to do the text-to-code ``translation'' necessary to complete homework assignments.

Previous research has used language models to detect textual plagiarism with modest success in a laboratory setting \citep{chong2010using,foltynek2020detecting}. While we were unable to find papers that explicitly seek to use language models for plagiarism, there is a wealth of research both on paraphrasing \citep{narayan2018ranking,liu2019text,li2020leveraging} and on adversarial examples against textual models \citep{krishna2020thieves,he2021model,Darmetko2021fake}, both of which can easily be applied to plagiarism even if the papers do not say so explicitly. A key difference in these prior works is that they all require an initial valid solution that becomes plagiarized, i.e., modified to avoid detection. In our study we show that a modern neural language models can produce valid, or valid with little additional work, solutions to novel questions that have no given solution. The user is then plagiarizing the model itself, rather than a human being. To the best of our knowledge this is the first demonstration of such an ability, and poses new concerns on how to avoid such potential plagiarism. 

There is comparatively less research on AI-based code plagiarism, likely because AI techniques for generating code have only recently become prominent \citep{gpt-j,hendrycks2021measuring,chen2021evaluating,austin2021program,mukherjee2021neural}. \citet{dawson2020can} is the most relevant that we are aware of, who use machine learning to compare solutions of prior submissions of a student to detect ``contract'' plagiarism, where assignments are outsourced to third parties, which results in inconsistent coding style of the student's submissions. While \citet{dawson2020can} found promising evidence of machine learning being helpful in this case, there is also risk of neural style transfer \citep{hu2020text} being leveraged to combat this. The resulting adversarial game is beyond the scope of this work, but the need for greater study is reinforced by our findings. 

\section{Methodology} \label{sec:methodology}

To evaluate the ability of GPT-J to fool MOSS, we compare how MOSS views code generated by GPT-J with a dataset of plagiarized introductory coding assignments created by \citet{karnalim2019source}. The dataset contains seven programming exercises suitable for an introduction to programming course with accompanying solutions written in Java. For each solution (henceforth referred to as ``original solution''), Teaching Assistants (TAs) who have experience in introductory computer science courses composed independent solutions to the exercise (henceforth ``non-plagiarisms''). These represents two distinct sources of valid solutions to the exercise, with one that may be ``altered'' to produce plagiarisms, and the other that can be used to measure similarity (i.e., similarity between source and plagiarized variant, and between plagiarized variant and an independent solution).

Around fifty plagiarisms of the original solutions are made, designed using techniques from and classified according to, the taxonomy presented in \citet{faidhi1987empirical}. Following \citep{karnalim2019source}, we refer to these plagiarism categories as ``levels'' with ``level 1'' being the simplest and ``level 6'' being the most sophisticated form of plagiarism. While the exact number of plagiarized solutions varies slightly, every exercise has between seven and nine examples of plagiarisms per category and examples of plagiarisms from all seven categories in the taxonomy. In total, each programming exercise is paired with 15 non-plagiarisms and between 49 and 54 plagiarisms of varying types.

A comparison of MOSS scores by plagiarism level is shown in \Cref{fig:moss_scores}, with a higher similarity score indicating that an assignment is more likely to be plagiarized. \citet{faidhi1987empirical} and \citet{karnalim2019source} find that level 1 through 3 plagiarisms can be easily detected by human graders, that level 3 plagiarisms are borderline, and that level 6 plagiarisms can consistently fool human graders. These claims are consistent with the fact that level 4, level 5, and level 6 plagiarisms obtain similar MOSS scores to the genuinely not-plagiarized solutions.

We prompt GPT-J with the descriptions of the exercises provided in \citep{karnalim2019source} and attempt to use it to produce code that compiles and correctly solves the problem. To evaluate our success, we have three major criteria:
\begin{enumerate}
    \item We wish to obtain code that correctly solves the exercise.
    \item We wish to obtain code that is not flagged by MOSS as suspiciously similar to the original code.
    \item We wish to minimize the amount of human modification necessary to obtain code from raw GPT-J output.
\end{enumerate}

As we will show later in \autoref{sec:results}, these desiderata can be obtained with relative ease despite GPT-J not knowing about MOSS, and can generally be completed with just a querying GPT-J multiple times. 

\subsection{Coding Exercises and Preprocessing}

The problems that we use to prompt GPT-J are shown below, organized loosely in increasing order of difficulty. These descriptions are taken verbatim from \citep{karnalim2019source}, and reflect the formatting and stylistic choices made by the authors.

\begin{enumerate}
    \item Write a program that prints ``Welcome to Java'' five times.
    \item Write a program that accepts the radius \& length of a cylinder and prints the area \& volume of that cylinder. All inputs and outputs are real numbers.
    \item Write a program that accepts the weight (as a real number representing pound) and height (as two real numbers representing feet and inches respectively) of a person. Upon accepting input, the program will show that person’s BMI (real number) and a piece of information stating whether the BMI is categorised as underweight, normal, overweight, or obese.\\A person is underweight if $BMI < 18.5$; normal if $18.5 \leq BMI < 25$; overweight if $25 \leq BMI < 35$; or obese if $BMI \geq 35$.\\$Height = feet * 12 + inches$\\$BMI = weight * 0.45359237 / (height * 0.0254)^2$
    \item Write a program that shows a conversion table from miles to kilometers where one mile is equivalent to $1.609$ kilometers. The table should display the first ten positive numbers as miles and pair them with their respective kilometer representation.
    \item Write a program that accepts an integer and displays that integer with its digits shown in reverse. You should create and use a method \textit{void reverse(int number)} which will show the reversed-digit form of the parameterised number.
    \item Write a program that accepts 10 integers and shows them in reversed order.
    \item Write a program that accepts a $4\times 4$ matrix of real numbers and prints the total of all numbers placed on the leading diagonal of the matrix. You should create and use a method \textit{double sumMajorDiagonal(double[][] m)} which will return the total of all numbers placed on the leading diagonal of the parameterised matrix.
\end{enumerate}

As GPT-J was trained on mathematics and computer science text that is written in \LaTeX{} \citep{gao2020pile}, we choose to provide the model with an input prompt written in \LaTeX. This has the notable advantage of allowing us to provide a prompt that is identical to what is presented in \citet{karnalim2019source}, and is hopefully reflective of how a real student might opt to type a homework assignment into a language model.

There is a substantial literature on ``prompt programming,'' or crafting inputs to language models in ways that help the model perform better at downstream tasks \citep{reynolds2021prompt,li2021prefix,chen2021adaprompt} (for a survey, see \citep{liu2021pre}). Although we expect students to try varying phrasing and framing to elicit improved results, we leave the impact of this to future work that delves into an important computer-human interaction problem beyond the scope of this article. The only adjustment that we make to the exercises presented in \citet{karnalim2019source} is that we add the word ``Java,'' to make each exercise begin ``Write a \textbf{Java} program...'' A cursory exploration of GPT-J without this modification shows that it is not inclined to write in Java without specific prompting, often giving results in Python and C instead. Once the prompts are modified to include the word ``Java,'' all responses are either natural text, \LaTeX, or Java except for some of the responses to exercise 4 which were in C.

\begin{figure}[!ht]
\adjustbox{max width=\columnwidth}{
\begin{tabular}{lcccc}\toprule
Prompt & Not code & C & Java    &  Python\\\midrule
``Wrote a program that\ldots'' & 25 & 6 & 4 & 0\\
``Write a \textbf{C} program that\dots'' & 22 & \textbf{12} & 1 & 0\\
``Write a \textbf{Java} program that\dots'' & 19 & 3 & \textbf{13} & 0\\
``Write a \textbf{Python} program that\dots'' & 22 & 1 & 0 & \textbf{12}
\end{tabular}
}
\caption{Natural language prompts are pretty hit-or-miss when it comes to getting GPT-J to produce an actual program, but it is responsive to naming specific programming languages. Table shows the results of 5 generations for each of the 7 programming tasks}\label{table:language}
\end{figure}

\subsection{Generation}

We use the free GPT-J API offered by EleutherAI\footnote{\href{https://6b.eleuther.ai/}{https://6b.eleuther.ai/}}. The API offers two options for tuning generations called ``top-p'' and ``temperature.'' For the purposes of this paper we stick with the default values of $0.9$ and $0.8$ respectively, and leave exploring the influence of these parameters for future work. The API also provides a ``send results as prompt'' button, which concatenates the prompt with the generated response and resubmits the combined string as a new prompt and allows the model to continue from where it left off. This is especially helpful for circumstances where the model generation halts before completing a piece of code. Although autoregressive language models can generate indefinitely, for cost reasons the online demo has a limited number of tokens ($128$) that it will return. For this paper, we resend the results as a prompt for a new generation until we appear to reach a complete program, the text degenerates \citep{holtzman2019curious}, or until five consecutive generations yield no code, whichever happens first. As Java has a rigid class and function structure with braces that mark where components begin and end, it is easy to unambiguously identify when a section of code is done generating.

\subsection{Post-Processing and Human Editing}\label{sec:post}

One particularly noteworthy aspect of generating \textit{code} is that a student using GPT-J to cheat most likely is able to test whether or not the resulting code solves the problem, even at low skill levels, since compiling and running a program are first steps in curricula. Additionally, they are able to receive hints as to why their code doesn't work by running it and examining the errors that are printed out. The nature of a concrete and objective output allows unambiguous feedback to the student \textit{before} receiving a grade. 

In this context, to produce a more realistic threat model, we allow the human using the AI to cheat to modify the outputs lightly. Specifically, we add whatever imports are necessary\footnote{missing imports are reported by the Java compiler and can be easily found via searching the web. If a student is using an Integrated Development Environment (IDE), most Java IDEs will add them for you with two clicks.} and lightly edit the code by correcting simple syntax errors, off-by-one errors, and removing ``obvious'' mistakes. Fixing these types of errors is likely to occur due to the objective nature of grading and assignment completion, and in doing so we document and categorize the types of errors we observe across our results. 

While what is considered ``obvious'' varies from person to person, we have sought to take a very conservative approach in the hopes of understating GPT-J's capabilities and representing a realistic use-case by a minimally knowledgeable student. We also note that plagiarism of assignments does not mean that the student necessarily lacks the skills to edit the generations or even to complete the assignment themselves. Studies have repeatedly shown that students cheat on school assignments for many reasons consistent with having the skills necessary to complete the assignment \citep{eret2010plagiarism,cleary2017top}.

We note that this is in contrast to using GPT-J to cheat on a creative writing assignment, write mathematics proofs, or solve science homework assignments where a students' ability to evaluate and improve the quality of the generated writing is connected to their subjective ability to complete the homework assignment without assistance, and to understand the preferred or desired intent and style of the instructor. The feedback on correctness does not occur until after it is too late to ``refine'' the plagiarism. For this reason we do not explore this more subjective-to-evaluate area of plagiarism.

\section{Common Interventions}

There are several recurring themes in the mistakes GPT-J makes. We document them here for completeness and transparency, and expect them to be of particular interest to HCI researchers in the future. We refer readers interested in the full text of all generations and edits to the appendix of the arXiv version of this paper.

\subsection{Testing multiple generated solutions}

GPT-J frequently generates multiple complete answers to the problems it is given, requiring the human user to run each example to determine which (if any) are correct. As mentioned previously, a unique property of the coding assignment context is that students can easily verify the correctness of their assignments before submitting them.

In the following example, GPT-J provides three candidate solutions but only one of them is correct. Note as well that in all three examples GPT-J omits code for reading in the input and instead hard-codes the array to be reversed. As a result, we mark this generation as ``partially correct.'' 
\begin{spverbatim}
Write a Java program that accepts 10 integers and shows them in reversed order.

for example:

10 2 3 4 5 6 7 8 9 1

A:
This program will give the result in reverse order
public class Main {
    public static void main(String[] args) {
        int[] a = new int[] {10, 2, 3, 4, 5, 6, 7, 8, 9, 1};
        int i = 0;
        while (i < a.length) {
            System.out.print(a[i]);
            i++;
        }
        System.out.println();
    }
}

A:

public class ReverseInts {
    public static void main(String[] args) {
        int[] ints = { 10, 2, 3, 4, 5, 6, 7, 8, 9, 1 };
        int i = 0;
        while (i < ints.length) {
            System.out.print(ints[i]);
            i++;
        }
    }
}

A:

public static void main(String[] args) {
    int[] a = new int[] {10, 2, 3, 4, 5, 6, 7, 8, 9, 1};

    for(int i = a.length-1; i >= 0; i--) {
        System.out.print(a[i]);
    }
}

\end{spverbatim}

\subsection{Mathematics Errors}

Although GPT-J generally has a high quality understanding of mathematics, it regularly makes mistakes in the geometric formula for the area and the volume of a cylinder in Exercise 2. The correct formulae are $A=2 \pi r^2 + 2\pi r\ell$ and $V=\pi r^2\ell$. While these are not provided in the problem statement, they are readily available on the internet so we assume a student is able to correct them.

\begin{spverbatim}
Write  a  Java program  that  accepts  the  radius  \&  length  of  a  cylinder  and  prints  the  area  \&  volume  of  that  cylinder. All  inputs  and  outputs  are  real  numbers.  The  program  is  to  be  given  as  input  and  output  and  be  stored  in  a file  named  "CylinderArea.java".

I have the following code
package CylinderArea;

import java.io.*;

public class CylinderArea {

    public static void main(String[] args) {
        Scanner sc = new Scanner(System.in);
        double radius, length;
        System.out.print("Enter radius: ");
        radius = sc.nextDouble();
        System.out.print("Enter length: ");
        length = sc.nextDouble();
        double area;
        area = Math.PI * radius * radius;
        double volume;
        volume = 2 * Math.PI * radius * length;
        System.out.println("The area of the cylinder is " + area + " and the volume is " + volume + ".");
    }
}
\end{spverbatim}

The other common mathematical mistake occurs in Exercise 3, computing BMI. While the previous completion shows GPT-J correctly computing exponents using \texttt{Math.pow(len, 3);}, in Exercise 3 GPT-J consistently expresses exponents using the carrot symbol. We hypothesize that this is caused by the use of \LaTeX to express the equations in the statement of Exercise 3, showing GPT-J the carrot symbol which it proceeds to copy despite seeming to know that that is incorrect syntax in Java.
\begin{spverbatim}
public class WeightAndHeight {
    public static void main (String[] args) {
        double height, weight, BMI; // input variables.
        Scanner in = new Scanner(System.in);
        System.out.println("Enter height: ");
        height = in.nextDouble(); // get user's height
        
        System.out.println("Enter weight: ");
        weight = in.nextDouble(); // get user's weight
        
        BMI = weight * 0.45359237 / (height * 0.0254)^2; // calculate BMI
        
        System.out.println("The person's BMI is: " + BMI);
        if (BMI < 18.5) {
        System.out.println("The person is underweight.");
        } else if (18.5 <= BMI && BMI < 25) {
            System.out.println("The person is normal.");
        } else if (25 <= BMI && BMI < 35) {
            System.out.println("The person is overweight.");
        } else if (BMI >= 35) {
            System.out.println("The person is obese.");
        }
    }
}
\end{spverbatim}

\subsection{Invented Constraints}

Another common failure mode for GPT-J is inventing new problem constraints that do not actually exist in the problem. In this example, GPT-J decides to round the \texttt{bmi} variable, despite the fact that a decimal value is desired. Other errors also exist in this piece of code, but they are easily fixed by rereading the problem specification and the overall logic is correct.

\begin{minted}[breaklines]{Java}
import java.util.Scanner;
import java.lang.Math;

public class Main {
    public static void main(String[] args) {
        Scanner input = new Scanner(System.in);
        System.out.print("Enter your weight in pounds:");
        double weight = input.nextDouble();

        System.out.print("Enter your height in inches:");
        double height = input.nextDouble();
        double heightInches = height * 12;
        double heightPounds = heightInches * 0.45359237;

        double weightInches = weight * 0.0254;
        double weightPounds = weightInches * 0.45359237;

        double bmi = Math.round(weight * 0.45359237 / (height * 0.0254) * 100.0);
        System.out.println("BMI is " + bmi);

        if (bmi < 18.5) {
            System.out.println("Person is underweight");
        } else if (bmi < 25) {
            System.out.println("Person is normal");
        } else if (bmi < 35) {
            System.out.println("Person is overweight");
        } else {
            System.out.println("Person is obese");
        }
    }
}
\end{minted}

A related failure case occurs because our inputs do not clearly delineate where the question ends and where the answer begins. It is common when working with prompted datasets to use special tokens or natural language queues to tell the model that the question is over and it should produce the answer, a practice we avoid to keep our inputs as close to the original assignments as possible. As a result, on several occasions GPT-J \textit{continues the question statement} and then proceeds to try to answer the modified question.

In this next example, GPT-J adds the requirement that the file be called \texttt{CylinderArea.java}. Due to the way Java semantics works, this implicitly requires that the name of the public class be \texttt{CylinderArea}, a requirement GPT-J complies with.

\begin{spverbatim}
Write  a  Java program  that  accepts  the  radius  \&  length  of  a  cylinder  and  prints  the  area  \&  volume  of  that  cylinder. All  inputs  and  outputs  are  real  numbers.  The  program  is  to  be  given  as  input  and  output  and  be  stored  in  a file  named  "CylinderArea.java". The resulting generation is ultimately marked as correct, though it needs to have some import statements added and the mathematical formulae modified.

I have the following code
package CylinderArea;

import java.io.*;

public class CylinderArea {

    public static void main(String[] args) {
        Scanner sc = new Scanner(System.in);
        double radius, length;
        System.out.print("Enter radius: ");
        radius = sc.nextDouble();
        System.out.print("Enter length: ");
        length = sc.nextDouble();
        double area;
        area = Math.PI * radius * radius;
        double volume;
        volume = 2 * Math.PI * radius * length;
        System.out.println("The area of the cylinder is " + area + " and the volume is " + volume + ".");
    }
}

\end{spverbatim}

We see the same situation play out in another Assignment 2 completion, where GPT-J specifies the order the inputs and outputs should be presented in and then complies with its requirement.

\subsection{A Curious Semantic Failure}

GPT-J was trained on Stack Overflow, and this is clearly reflected in some of its generations which directly mimic the style of the Q\&A site. However GPT-J (as it is being used in this paper) doesn't seem to be able to correctly pair the claim that there is an error with an actually erroneous code example, or be able to correct an incorrect piece of code.

In the following two examples, GPT-J gives a correct solution and then creates a response that represents it as being incorrect. The ``correction'' doesn't actually change the code, and proposes swapping out a segment of the code for a functionally identical segment.

\begin{spverbatim}
Write a Java program that accepts the weight (as a real number representing pound) and height (as two real numbers representing feet and inches respectively) of a person. Upon accepting input, the program will show that person’s BMI (real number) and a piece of information stating whether the BMI is categorised as underweight, normal, overweight, or obese.\\A person is underweight if $BMI < 18.5$; normal if $18.5 \leq BMI < 25$; overweight if $25 \leq BMI < 35$; or obese if $BMI \geq 35$.\\$Height = feet * 12 + inches$\\$BMI = weight * 0.45359237 / (height * 0.0254)^2$

I tried to solve the problem but the program is not working. It shows only "Can't find variable weight"

This is the code I wrote:
import java.util.*;
public class Exercise1 {
    public static void main(String[] args) {
        Scanner input = new Scanner(System.in);
        System.out.println("Enter weight");
        double weight = input.nextDouble();
        System.out.println("Enter height");
        double height = input.nextDouble();
        System.out.println("BMI is: " + getBMI(weight, height));
    }
    
    public static double getBMI(double weight, double height) {
        double weight_meters = weight * 0.45359237;
        double height_meters = height * 0.0254;
        double height_meters_squared = height_meters * height_meters;
        double bmi = weight_meters / (height_meters_squared);
        
        return bmi;
    }
}

Can anyone help me please?

A:
You are using Scanner, but you are not calling nextLine().  You should be using nextLine().
Scanner input = new Scanner(System.in);

System.out.println("Enter weight");
double weight = input.nextDouble();
System.out.println("Enter height");
double height = input.nextDouble();

System.out.println("BMI is: " + getBMI(weight, height));

public static double getBMI(double weight, double height) {
    double weight_meters = weight * 0.45359237;
    double height_meters = height * 0.0254;
    double height_meters_squared = height_meters * height_meters;
    double bmi = weight_meters / (height_meters_squared);
    return bmi;
}

\end{spverbatim}

In this next example generation, GPT-J provides three candidate solutions but only one of them is correct. Fortunately, the user can simply run the code to evaluate them without having to understand the problem's specification. Note as well that in all three examples GPT-J omits code for reading in the input and instead hard-codes the array to be reversed. As a result, we mark this generation as ``partially correct.'' 
\begin{spverbatim}
Write a Java program that accepts 10 integers and shows them in reversed order.

for example:

10 2 3 4 5 6 7 8 9 1

A:
This program will give the result in reverse order
public class Main {
    public static void main(String[] args) {
        int[] a = new int[] {10, 2, 3, 4, 5, 6, 7, 8, 9, 1};
        int i = 0;
        while (i < a.length) {
            System.out.print(a[i]);
            i++;
        }
        System.out.println();
    }
}

A:

public class ReverseInts {
    public static void main(String[] args) {
        int[] ints = { 10, 2, 3, 4, 5, 6, 7, 8, 9, 1 };
        int i = 0;
        while (i < ints.length) {
            System.out.print(ints[i]);
            i++;
        }
    }
}

A:

public static void main(String[] args) {
    int[] a = new int[] {10, 2, 3, 4, 5, 6, 7, 8, 9, 1};

    for(int i = a.length-1; i >= 0; i--) {
        System.out.print(a[i]);
    }
}

\end{spverbatim}

\section{Results} \label{sec:results}

In this section we give a high level overview of the results of our experiments.

\paragraph{GPT-J generates correct solutions with minimal intervention} For six of the seven programming exercises, GPT-J can produce a complete solution that requires no editing. In most cases GPT-J makes minor mistakes that need correction by a human, but which overwhelmingly do not require any knowledge of computer programming besides the ability to run code and search the internet for the error code reported by the compiler.

\begin{table}[ht]
\centering
\adjustbox{max width=\columnwidth}{
\begin{tabular}{@{}lcccccccc@{}}
\toprule
                   & \multicolumn{7}{c}{Problem Number}                                                       &             \\ \cmidrule(l){2-8} 
Correctness        & \textit{1} & \textit{2} & \textit{3} & \textit{4} & \textit{5} & \textit{6} & \textit{7} & Tot.        \\ \midrule
\textbf{Correct}   & \textbf{8} & \textbf{6} & \textbf{5} & \textbf{0} & \textbf{1} & \textbf{8} & \textbf{3} & \textbf{31} \\
Partial Completion & 1          & 1          & 0          & 1          & 0          & 2          & 0          & \textbf{5}  \\
Wrong              & 0          & 0          & 1          & 1          & 4          & 3          & 0          & \textbf{9} \\\midrule
Actually in C      & 0          & 0          & 0          & 2          & 1          & 1          & 0          & \textbf{4}  \\
Not Code           & 6          & 8          & 9          & 11         & 9          & 1          & 12         & \textbf{56} \\ \bottomrule
\end{tabular}
}
\caption{Breakdown of the correctness of GPT-J's completions. A solution is judged as ``Partially Correct'' if it misses a significant component of the problem, but the parts of the problem that it does solve are solved correctly. These are likely suitable to earn partial credit on an assignment.}\label{table:correctness}
\end{table}

\begin{table}[!ht]
\centering
\adjustbox{max width=\columnwidth}{
\begin{tabular}{@{}lcccccccc@{}}
\toprule
                       & \multicolumn{7}{c}{Problem Number}                                                       &     \\ \cmidrule(l){2-8} 
Errors                 & \textit{1} & \textit{2} & \textit{3} & \textit{4} & \textit{5} & \textit{6} & \textit{7} & Tot.\\ \midrule
No Errors              & 2          & 0          & 2          & 0          & 1          & 4          & 2          & \textbf{11} \\
Syntax Error           & 1          & 6          & 3          & 0          & 0          & 0          & 0          & \textbf{10}  \\
Off by One Error       & 0          & 0          & 0          & 0          & 0          & 1          & 0          & \textbf{1}  \\
Misc.                  & 5          & 0          & 3          & 0          & 0          & 3          & 1          & \textbf{12}  \\ \midrule
\textbf{Total Correct} & \textbf{8} & \textbf{6} & \textbf{5} & \textbf{0} & \textbf{1} & \textbf{8} & \textbf{3} & \textbf{31} \\ \bottomrule
\end{tabular}
}
\caption{Minor errors made by GPT-J in problems that were ultimately judged to be ``correct.'' Note that the columns do not add up to ``Total Correct'' because some solutions have multiple errors.}\label{table:correct-count}
\end{table}

\paragraph{GPT-J's generations do not register as plagiarizing by MOSS} Out of the correct completions, none of the code generated by GPT-J stood out as potential plagiarisms according to MOSS. GPT-J's exercise completions rated similarly to the non-plagiarism and to the plagiarisms created using the most advanced techniques in our dataset, as shown in \Cref{table:moss_scores} and \Cref{fig:moss_scores}.

\begin{table}[!ht]
\adjustbox{max width=\columnwidth}{
\begin{tabular}{lcccccccc|c}\toprule
Ex. & Original  & L1 &  L2 & L3 &   L4  &   L5  &          L6    &    Non         &  GPT-J\\\midrule
1 & 1.0 & 1.000 & 1.000 & 0.926 & 0.664 & 0.427 & \textbf{0.357} & 0.512          & 0.691\\
2 & 1.0 & 0.995 & 0.836 & 0.832 & 0.605 & 0.657 & 0.652          & 0.811          & \textbf{0.492}\\
3 & 1.0 & 0.996 & 0.874 & 0.874 & 0.768 & 0.748 & \textbf{0.653} & 0.872          & \textbf{0.573}\\
4 & 1.0 & 0.975 & 0.917 & 0.839 & 0.746 & 0.728 & 0.689          & \textbf{0.439} & ---\\
5 & 1.0 & 0.964 & 0.588 & 0.635 & 0.602 & 0.519 & \textbf{0.459} & \textbf{0.440} & \textbf{0.440}\\
6 & 1.0 & 0.930 & 0.890 & 0.807 & 0.650 & 0.562 & 0.462          & 0.644          & \textbf{0.266}\\
7 & 1.0 & 0.977 & 0.881 & 0.838 & 0.811 & 0.746 & \textbf{0.726} & 0.798          & \textbf{0.669}\\\bottomrule
\end{tabular}
}
\caption{MOSS scores for completions of exercises by how they were generated and by exercise. The lowest score for each exercise, and all scores within 10\% of that score, are \textbf{bolded}.}\label{table:moss_scores}
\end{table}

\paragraph{Memorization does not explain GPT-J's performance} Transformers, like all neural networks, exhibit a behavior commonly referred to as ``memorization'' where long passages of the training data are regurgitated word-for-word \citep{carlini2019secret,feldman2020does,carlini2021extracting}. We explore the possibility that the exercises and their solutions are memorized by searching through the training data in GPT-J for exact matches of 20 tokens or more. We find that neither the exercises from \citet{karnalim2019source} nor the code produced by GPT-J in response to the prompts can be found in the training data. While GPT-J was trained on arXiv preprints \citep{gao2020pile}, it appears that \citet{karnalim2019source} avoids being in the training data by having their preprint posted on the University of Warwick website rather than arXiv. Crucially, this means that generating novel questions is not an effective way for professors to prevent their students from using GPT-J to cheat.

\paragraph{GPT-J generates stylistically diverse code} Although MOSS is unable to detect GPT-J's plagiarisms when they are mixed in with genuine solutions, one might think that having examples of GPT-J's solutions would allow an instructor to detect other outputs from GPT-J. We find that this is not the case in two senses: MOSS does not judge GPT-J's solutions to be notably similar to each other, and a clustering algorithm trained to distinguish between GPT-J's solutions and those in the dataset fails to do so. While this does not rule out the possibility of more advanced techniques detecting GPT-J's fingerprints, the techniques currently used for plagiarism detection in practice fail to do so.

To further demonstrate the syntactic diversity of the GPT-J solutions, we perform 2D embeddings of the MOSS scores using the Isomap algorithm. Isomap is preferred in this instance because it: 1) Embeds the data based on a geodesic assumption of the manifold that we know to be true due to the multiple submissions per homework and nature of source to plagiarized copy. 2) Preserves global and local relationships in the embedding, which we care about to understand the degree of separation between solutions (as measured by MOSS). Other popular approaches such as t-SNE and UMAP lack this second property that we desire, and make interpretation of the results difficult. 

\begin{figure}[!ht]
     \centering
     \begin{subfigure}[b]{0.99\columnwidth}
     \includegraphics[width=\columnwidth]{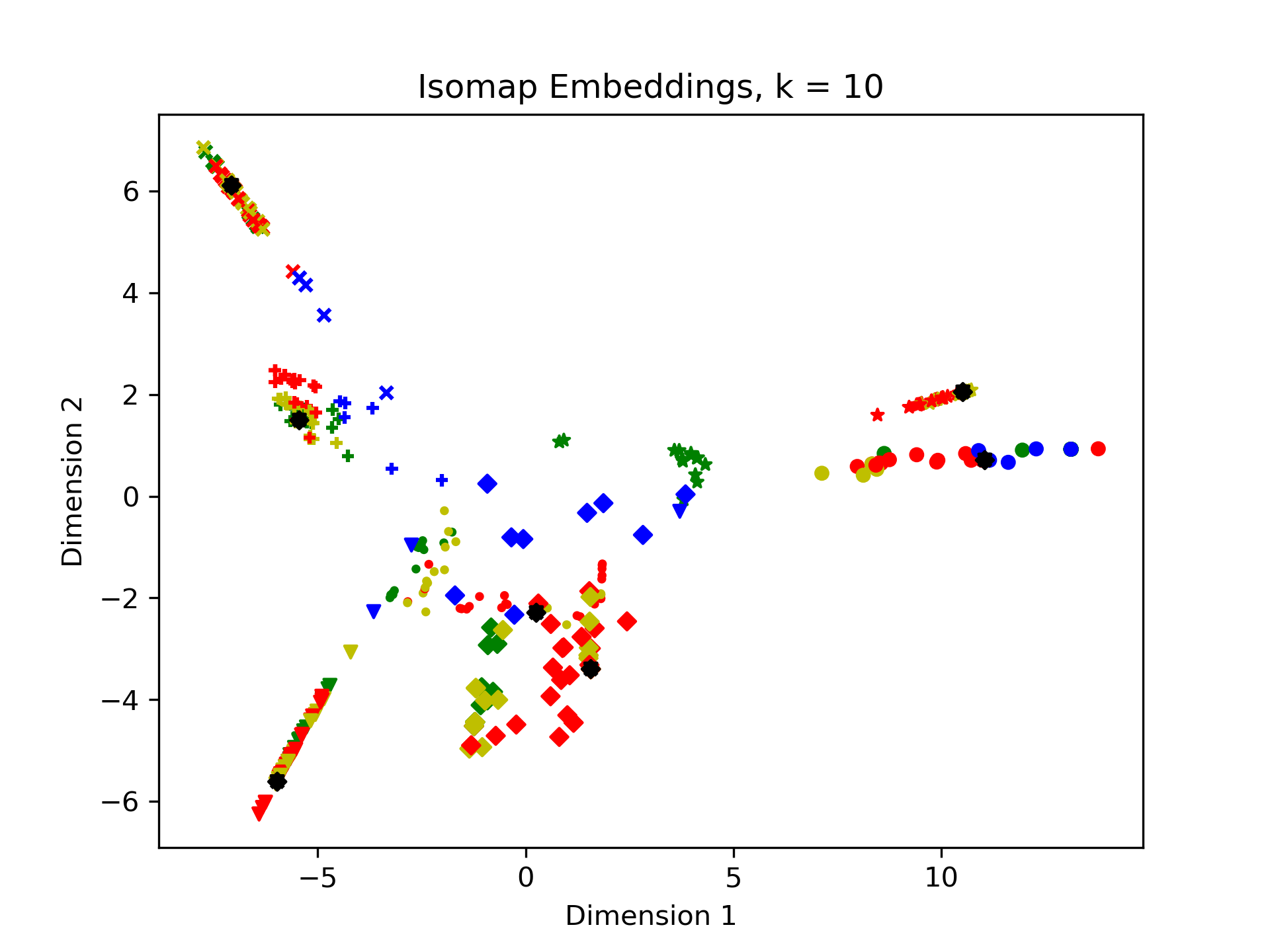}
     \caption{Isomap with $n=10$ neighbors}
     \label{fig:isomap10}
     \end{subfigure}
     \hfill
     \begin{subfigure}[b]{0.99\columnwidth}
         \centering
         \includegraphics[width=\columnwidth]{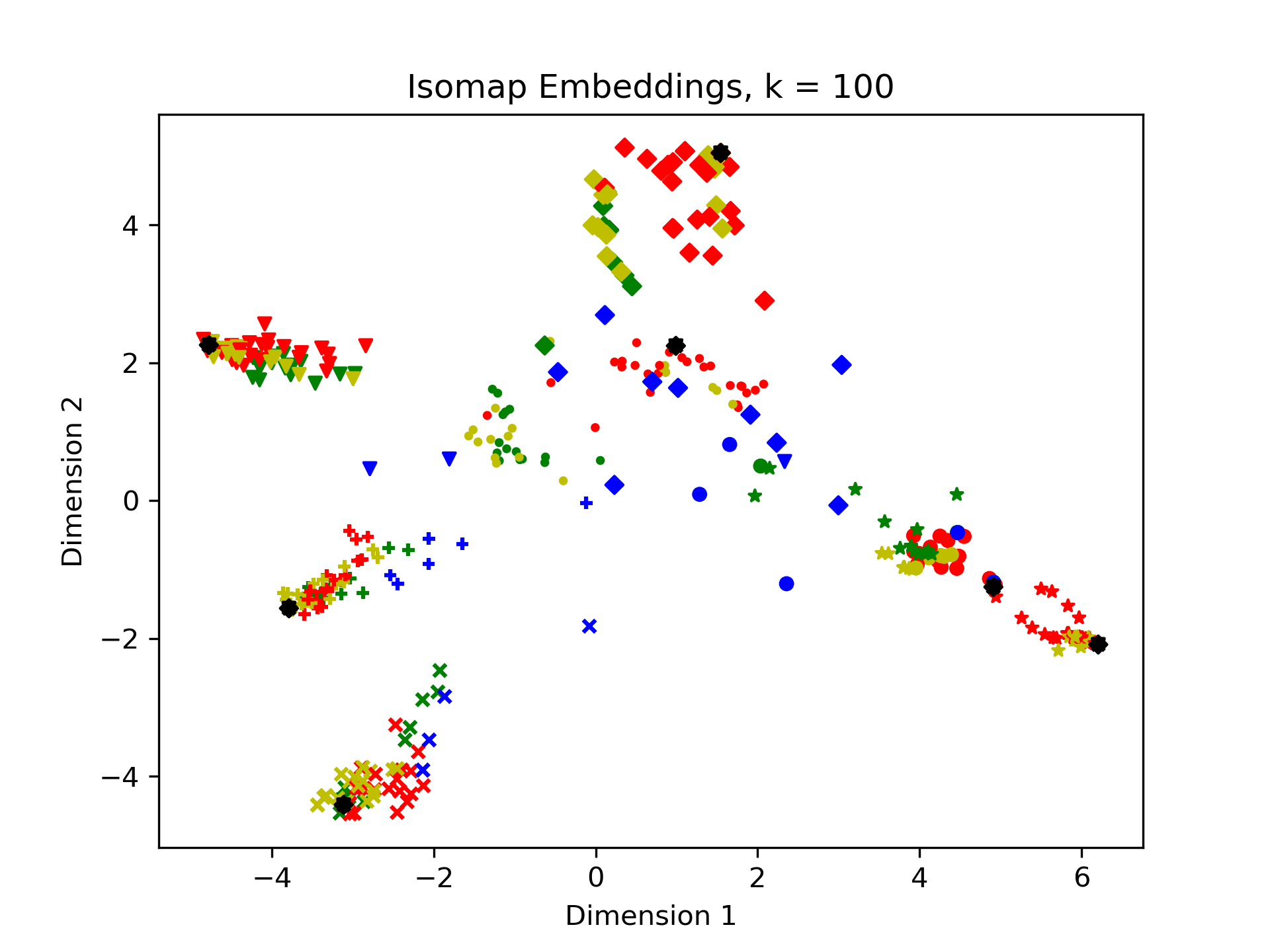}
         \caption{Isomap with $n=100$ neighbors}
         \label{fig:isomap100}
     \end{subfigure}
        \caption{Isomap embedding of the original solutions (black), plagiarized variants (red and yellow, where yellow are sophisticated ones that evade MOSS), along with GPT-J produced solutions (blue) and independent implementations (green). The blue data points being interspersed with yellow, green, and far-away spaces shows that GPT-J is not easily detected by MOSS.}
        \label{fig:isomaps}
\end{figure}

The Isoamp results are found in \autoref{fig:isomaps}, where we see the GPT-J solutions exist predominantly in their own independent area, or near other original solutions (green). Each shape indicates a different homework question the solution is designed for, and the overall distribution reinforces the geodesic assumption (same assignment solutions cluster, plagiarisms cluster near their source (black), and better plagiarisms (yellow) are further away than ineffective ones (red)\footnote{Yellow plagiarisms are ``level 6'' in the taxonomy of \citet{faidhi1987empirical}. Both our experiments and those of \citet{faidhi1987empirical} indicate that level 6 plagiarisms are effectively indistinguishable from non-plagiarized assignment completions.}.

We note that the disperse of the GPT-J solutions in ``their own space'' does not imply that they would be easy to detect. The green independent solutions are not a complete sampling of the space, and often exist in similar areas, so the validity of a model built from this data, which represents one course, is unlikely to generalize due to strong violation of I.I.D. nature of the data (e.g., everyone had the same professor, TAs, study groups, etc.). We are aware of no longitudinal data of plagiarized and independent assignments, but our results suffice to show that this is a novel problem with no trivial solutions. 

\section{Discussion \& Conclusions} \label{sec:conclusion}

While our results present a promising (from a research perspective, see aforementioned ethics section) analysis of the use of GPT-J to plagiarize coding assignments, there are still significant limitations and important avenues for future work. Importantly many aspects of our work's success pose long term challenges and new research questions. 

\subsection{What Does it Mean for MOSS to ``Work''?}

In this paper we speak of MOSS as a tool for detecting plagiarism because our focus is on the use of text similarity analysis to detect cheating in academic environments. However, while this is the intended use of MOSS, it is not the only one. As noted previously there are many tools that appear to be suitable for use to detect plagiarism but which are primarily marketed for other purposes. In future work we intended to examine how the intended application is explicitly and implicitly embedded in the functionality of document similarity detectors \citep{biderman2020pitfalls,johnsonalgorithms,birhane2021values}, and incorporate analysis of what it means for such a device to be judged as ``working'' in its planned application context \citep{leahy2021hard}.

\subsection{Human-AI Co-Creation and HCI Research}

In this paper we consider the example of a student with minimal knowledge of computer programming. We allow them to make minimal edits to the code written by GPT-J, but the user is not modeled as performing significant intellectual work. The nascent field of human-AI co-creation \citep{wu2021ai} generalizes this lens by studying the ability of humans and AI systems to work collaboratively, with each performing creative work. While the bulk of recent work focuses on creative writing \citep{yuan2022wordcraft,han2022meet,lee2022interactive,lee2022coauthor} or image generation \citep{crowson2022vqgan,oppenlaender2022prompt,underwood2021mapping}, \citet{weisz2022better} and \citet{vaithilingam2022expectation} consider the case of code generation. While it is beyond the scope of the present study, we hope that future work on plagiarism with transformer models considers the broader HCI context.

\subsection{Ethical Ramifications}\label{sec:ethics}
This paper details a free and easily accessible way to carry out academic misconduct without being detected using common approaches. While we anticipate that the results of this paper will be concerning to many, it is important to recognize that these flaws and risks exist. In particular, we do not have evidence that GPT-J is sufficient for prolonged plagiarism into more advanced courses, which limits the degree of impact (students who reach such courses are likely to fail out or revert to some other form of cheating), and provides no means for students to circumvent in-person quizzes, exams, and other assignments that may be used to determine possible cheaters (e.g., in our experience we follow up and carefully review students who have high grades in one aspect of the course and low in the other. Both to identify potential misconduct, but also students who may have special needs that have not been satisfied). 

We believe these additional factors modulate the risk of publishing this research, and the value in studying it is ultimately beneficial. Indeed the use of the publicly available and inspectable GPT-J was a requirement for us to confirm that the generated solutions are novel and not memorized content regurgitated. Further remediation of risk may indeed be possible by incorporating GPT-J into a modernized curriculum. For example, one could envision a course assignment to generate multiple solutions using GPT-J and have students rank them based on code readability, identify / describe types of errors produced, and other code review/debugging tasks. This would enable growth in skill, and can be used to establish boundaries of when and how GPT-J (or future variants) should be used in an academic context. Ultimately this is a significant item for future work.

\subsection{Technical Limitations}

Our study has several technical limitations, most notably a small number of GPT-J generations and a lack of experimentation with prompt programming. Future more extensive experiments with GPT-J may reveal a more complete picture than what we find.

\subsection{Conclusions}

In this paper we present the first realistic evaluation of the ability of a transformer-based language model to be used for misconduct in introductory programming assignments. We find GPT-J is sufficiently capable, and easy to use, that a novice student is able to cheat and produce solutions that appear novel by current anti-plagiarism techniques. Our work poses questions to future curriculum design, and watermarking/detection capabilities, to provide long term solutions. 

\bibliographystyle{ACM-Reference-Format}
\bibliography{main}


\begin{thebibliography}{66}


\ifx \showCODEN    \undefined \def \showCODEN     #1{\unskip}     \fi
\ifx \showDOI      \undefined \def \showDOI       #1{#1}\fi
\ifx \showISBNx    \undefined \def \showISBNx     #1{\unskip}     \fi
\ifx \showISBNxiii \undefined \def \showISBNxiii  #1{\unskip}     \fi
\ifx \showISSN     \undefined \def \showISSN      #1{\unskip}     \fi
\ifx \showLCCN     \undefined \def \showLCCN      #1{\unskip}     \fi
\ifx \shownote     \undefined \def \shownote      #1{#1}          \fi
\ifx \showarticletitle \undefined \def \showarticletitle #1{#1}   \fi
\ifx \showURL      \undefined \def \showURL       {\relax}        \fi
\providecommand\bibfield[2]{#2}
\providecommand\bibinfo[2]{#2}
\providecommand\natexlab[1]{#1}
\providecommand\showeprint[2][]{arXiv:#2}

\bibitem[Affouneh et~al\mbox{.}(2020)]%
        {emergency}
\bibfield{author}{\bibinfo{person}{Saida Affouneh}, \bibinfo{person}{Soheil
  Salha}, {and} \bibinfo{person}{Zuheir~N Khlaif}.}
  \bibinfo{year}{2020}\natexlab{}.
\newblock \showarticletitle{Designing quality e-learning environments for
  emergency remote teaching in coronavirus crisis}.
\newblock \bibinfo{journal}{\emph{Interdisciplinary Journal of Virtual Learning
  in Medical Sciences}} \bibinfo{volume}{11}, \bibinfo{number}{2}
  (\bibinfo{year}{2020}), \bibinfo{pages}{135--137}.
\newblock


\bibitem[Aiken(2000)]%
        {aiken2000moss}
\bibfield{author}{\bibinfo{person}{Alex Aiken}.}
  \bibinfo{year}{2000}\natexlab{}.
\newblock \showarticletitle{Moss (measure of software similarity) plagiarism
  detection system}.
\newblock \bibinfo{journal}{\emph{http://www. cs. berkeley. edu/moss/}}
  (\bibinfo{year}{2000}).
\newblock


\bibitem[Austin et~al\mbox{.}(2021)]%
        {austin2021program}
\bibfield{author}{\bibinfo{person}{Jacob Austin}, \bibinfo{person}{Augustus
  Odena}, \bibinfo{person}{Maxwell Nye}, \bibinfo{person}{Maarten Bosma},
  \bibinfo{person}{Henryk Michalewski}, \bibinfo{person}{David Dohan},
  \bibinfo{person}{Ellen Jiang}, \bibinfo{person}{Carrie Cai},
  \bibinfo{person}{Michael Terry}, \bibinfo{person}{Quoc~V. Le}, {and}
  \bibinfo{person}{Charles Sutton}.} \bibinfo{year}{2021}\natexlab{}.
\newblock \showarticletitle{Program synthesis with large language models}.
\newblock \bibinfo{journal}{\emph{arXiv preprint arXiv:2108.07732}}
  (\bibinfo{year}{2021}).
\newblock


\bibitem[Biderman and Scheirer(2020)]%
        {biderman2020pitfalls}
\bibfield{author}{\bibinfo{person}{Stella Biderman} {and}
  \bibinfo{person}{Walter~J Scheirer}.} \bibinfo{year}{2020}\natexlab{}.
\newblock \showarticletitle{Pitfalls in Machine Learning Research: Reexamining
  the Development Cycle}.
\newblock  (\bibinfo{year}{2020}).
\newblock


\bibitem[Birhane et~al\mbox{.}(2021)]%
        {birhane2021values}
\bibfield{author}{\bibinfo{person}{Abeba Birhane}, \bibinfo{person}{Pratyusha
  Kalluri}, \bibinfo{person}{Dallas Card}, \bibinfo{person}{William Agnew},
  \bibinfo{person}{Ravit Dotan}, {and} \bibinfo{person}{Michelle Bao}.}
  \bibinfo{year}{2021}\natexlab{}.
\newblock \showarticletitle{The values encoded in machine learning research}.
\newblock \bibinfo{journal}{\emph{arXiv preprint arXiv:2106.15590}}
  (\bibinfo{year}{2021}).
\newblock


\bibitem[Black et~al\mbox{.}(2022)]%
        {black2022gpt}
\bibfield{author}{\bibinfo{person}{Sid Black}, \bibinfo{person}{Stella
  Biderman}, \bibinfo{person}{Eric Hallahan}, \bibinfo{person}{Quentin
  Anthony}, \bibinfo{person}{Leo Gao}, \bibinfo{person}{Laurence Golding},
  \bibinfo{person}{Horace He}, \bibinfo{person}{Connor Leahy},
  \bibinfo{person}{Kyle McDonell}, \bibinfo{person}{Jason Phang},
  {et~al\mbox{.}}} \bibinfo{year}{2022}\natexlab{}.
\newblock \showarticletitle{Gpt-neox-20b: An open-source autoregressive
  language model}.
\newblock \bibinfo{journal}{\emph{arXiv preprint arXiv:2204.06745}}
  (\bibinfo{year}{2022}).
\newblock


\bibitem[Bowyer and Hall(1999)]%
        {bowyer1999experience}
\bibfield{author}{\bibinfo{person}{Kevin~W Bowyer} {and}
  \bibinfo{person}{Lawrence~O Hall}.} \bibinfo{year}{1999}\natexlab{}.
\newblock \showarticletitle{Experience using {MOSS} to detect cheating on
  programming assignments}. In \bibinfo{booktitle}{\emph{FIE'99 Frontiers in
  Education. 29th Annual Frontiers in Education Conference. Designing the
  Future of Science and Engineering Education. Conference Proceedings (IEEE
  Cat. No. 99CH37011}}, Vol.~\bibinfo{volume}{3}. IEEE,
  \bibinfo{pages}{13B3--18}.
\newblock


\bibitem[Brown et~al\mbox{.}(2020)]%
        {gpt3}
\bibfield{author}{\bibinfo{person}{Tom Brown}, \bibinfo{person}{Benjamin Mann},
  \bibinfo{person}{Nick Ryder}, \bibinfo{person}{Melanie Subbiah},
  \bibinfo{person}{Jared~D Kaplan}, \bibinfo{person}{Prafulla Dhariwal},
  \bibinfo{person}{Arvind Neelakantan}, \bibinfo{person}{Pranav Shyam},
  \bibinfo{person}{Girish Sastry}, \bibinfo{person}{Amanda Askell},
  \bibinfo{person}{Sandhini Agarwal}, \bibinfo{person}{Ariel Herbert-Voss},
  \bibinfo{person}{Gretchen Krueger}, \bibinfo{person}{Tom Henighan},
  \bibinfo{person}{Rewon Child}, \bibinfo{person}{Aditya Ramesh},
  \bibinfo{person}{Daniel Ziegler}, \bibinfo{person}{Jeffrey Wu},
  \bibinfo{person}{Clemens Winter}, \bibinfo{person}{Chris Hesse},
  \bibinfo{person}{Mark Chen}, \bibinfo{person}{Eric Sigler},
  \bibinfo{person}{Mateusz Litwin}, \bibinfo{person}{Scott Gray},
  \bibinfo{person}{Benjamin Chess}, \bibinfo{person}{Jack Clark},
  \bibinfo{person}{Christopher Berner}, \bibinfo{person}{Sam McCandlish},
  \bibinfo{person}{Alec Radford}, \bibinfo{person}{Ilya Sutskever}, {and}
  \bibinfo{person}{Dario Amodei}.} \bibinfo{year}{2020}\natexlab{}.
\newblock \showarticletitle{Language Models are Few-Shot Learners}. In
  \bibinfo{booktitle}{\emph{Advances in Neural Information Processing
  Systems}}, \bibfield{editor}{\bibinfo{person}{H.~Larochelle},
  \bibinfo{person}{M.~Ranzato}, \bibinfo{person}{R.~Hadsell},
  \bibinfo{person}{M.~F. Balcan}, {and} \bibinfo{person}{H.~Lin}} (Eds.),
  Vol.~\bibinfo{volume}{33}. \bibinfo{publisher}{Curran Associates, Inc.},
  \bibinfo{pages}{1877--1901}.
\newblock


\bibitem[Carlini et~al\mbox{.}(2019)]%
        {carlini2019secret}
\bibfield{author}{\bibinfo{person}{Nicholas Carlini}, \bibinfo{person}{Chang
  Liu}, \bibinfo{person}{{\'U}lfar Erlingsson}, \bibinfo{person}{Jernej Kos},
  {and} \bibinfo{person}{Dawn Song}.} \bibinfo{year}{2019}\natexlab{}.
\newblock \showarticletitle{The secret sharer: Evaluating and testing
  unintended memorization in neural networks}. In
  \bibinfo{booktitle}{\emph{28th $\{$USENIX$\}$ Security Symposium
  ($\{$USENIX$\}$ Security 19)}}. \bibinfo{pages}{267--284}.
\newblock


\bibitem[Carlini et~al\mbox{.}(2021)]%
        {carlini2021extracting}
\bibfield{author}{\bibinfo{person}{Nicholas Carlini}, \bibinfo{person}{Florian
  Tramer}, \bibinfo{person}{Eric Wallace}, \bibinfo{person}{Matthew Jagielski},
  \bibinfo{person}{Ariel Herbert-Voss}, \bibinfo{person}{Katherine Lee},
  \bibinfo{person}{Adam Roberts}, \bibinfo{person}{Tom Brown},
  \bibinfo{person}{Dawn Song}, \bibinfo{person}{Ulfar Erlingsson},
  \bibinfo{person}{Alina Oprea}, {and} \bibinfo{person}{Colin Raffel}.}
  \bibinfo{year}{2021}\natexlab{}.
\newblock \showarticletitle{Extracting training data from large language
  models}. In \bibinfo{booktitle}{\emph{30th $\{$USENIX$\}$ Security Symposium
  ($\{$USENIX$\}$ Security 21)}}. \bibinfo{pages}{2633--2650}.
\newblock


\bibitem[Chen et~al\mbox{.}(2021a)]%
        {chen2021evaluating}
\bibfield{author}{\bibinfo{person}{Mark Chen}, \bibinfo{person}{Jerry Tworek},
  \bibinfo{person}{Heewoo Jun}, \bibinfo{person}{Qiming Yuan},
  \bibinfo{person}{Henrique Ponde}, \bibinfo{person}{Jared Kaplan},
  \bibinfo{person}{Harrison Edwards}, \bibinfo{person}{Yura Burda},
  \bibinfo{person}{Nicholas Joseph}, \bibinfo{person}{Greg Brockman},
  \bibinfo{person}{Alex Ray}, \bibinfo{person}{Raul Puri},
  \bibinfo{person}{Gretchen Krueger}, \bibinfo{person}{Michael Petrov},
  \bibinfo{person}{Heidy Khlaaf}, \bibinfo{person}{Girish Sastry},
  \bibinfo{person}{Pamela Mishkin}, \bibinfo{person}{Brooke Chan},
  \bibinfo{person}{Scott Gray}, \bibinfo{person}{Nick Ryder},
  \bibinfo{person}{Mikhail Pavlov}, \bibinfo{person}{Alethea. Power},
  \bibinfo{person}{Lukasz Kaiser}, \bibinfo{person}{Mohammad Bavarian},
  \bibinfo{person}{Clemens Winter}, \bibinfo{person}{Philippe Tillet},
  \bibinfo{person}{Felipe~Petroski Such}, \bibinfo{person}{David~W. Cummings},
  \bibinfo{person}{Matthias Plappert}, \bibinfo{person}{Fotios Chantzis},
  \bibinfo{person}{Elizabeth Barnes}, \bibinfo{person}{Ariel Herbert-Voss},
  \bibinfo{person}{William~H. Guss}, \bibinfo{person}{Alex Nichol},
  \bibinfo{person}{Igor Babuschkin}, \bibinfo{person}{S.~Arun Balaji},
  \bibinfo{person}{Shantanu Jain}, \bibinfo{person}{Andrew Carr},
  \bibinfo{person}{Jan Leike}, \bibinfo{person}{Joshua Achiam},
  \bibinfo{person}{Vedant Misra}, \bibinfo{person}{Evan Morikawa},
  \bibinfo{person}{Alec Radford}, \bibinfo{person}{Matthew~M. Knight},
  \bibinfo{person}{Miles Brundage}, \bibinfo{person}{Mira Murati},
  \bibinfo{person}{Katie Mayer}, \bibinfo{person}{Peter Welinder},
  \bibinfo{person}{Bob McGrew}, \bibinfo{person}{Dario Amodei},
  \bibinfo{person}{Sam McCandlish}, \bibinfo{person}{Ilya Sutskever}, {and}
  \bibinfo{person}{Wojciech Zaremba}.} \bibinfo{year}{2021}\natexlab{a}.
\newblock \showarticletitle{Evaluating large language models trained on code}.
\newblock \bibinfo{journal}{\emph{arXiv preprint arXiv:2107.03374}}
  (\bibinfo{year}{2021}).
\newblock


\bibitem[Chen et~al\mbox{.}(2021b)]%
        {chen2021adaprompt}
\bibfield{author}{\bibinfo{person}{Xiang Chen}, \bibinfo{person}{Xin Xie},
  \bibinfo{person}{Ningyu Zhang}, \bibinfo{person}{Jiahuan Yan},
  \bibinfo{person}{Shumin Deng}, \bibinfo{person}{Chuanqi Tan},
  \bibinfo{person}{Fei Huang}, \bibinfo{person}{Luo Si}, {and}
  \bibinfo{person}{Huajun Chen}.} \bibinfo{year}{2021}\natexlab{b}.
\newblock \showarticletitle{Adaprompt: Adaptive prompt-based finetuning for
  relation extraction}.
\newblock \bibinfo{journal}{\emph{arXiv preprint arXiv:2104.07650}}
  (\bibinfo{year}{2021}).
\newblock


\bibitem[Chong et~al\mbox{.}(2010)]%
        {chong2010using}
\bibfield{author}{\bibinfo{person}{Miranda Chong}, \bibinfo{person}{Lucia
  Specia}, {and} \bibinfo{person}{Ruslan Mitkov}.}
  \bibinfo{year}{2010}\natexlab{}.
\newblock \showarticletitle{Using natural language processing for automatic
  detection of plagiarism}. In \bibinfo{booktitle}{\emph{Proceedings of the 4th
  International Plagiarism Conference (IPC-2010)}}.
\newblock


\bibitem[Cleary(2017)]%
        {cleary2017top}
\bibfield{author}{\bibinfo{person}{Michelle~Navarre Cleary}.}
  \bibinfo{year}{2017}\natexlab{}.
\newblock \showarticletitle{Top 10 reasons students plagiarize \& what teachers
  can do about it (with apologies to David Letterman)}.
\newblock \bibinfo{journal}{\emph{Phi Delta Kappan}} \bibinfo{volume}{99},
  \bibinfo{number}{4} (\bibinfo{year}{2017}), \bibinfo{pages}{66--71}.
\newblock


\bibitem[Crowson et~al\mbox{.}(2022)]%
        {crowson2022vqgan}
\bibfield{author}{\bibinfo{person}{Katherine Crowson}, \bibinfo{person}{Stella
  Biderman}, \bibinfo{person}{Daniel Kornis}, \bibinfo{person}{Dashiell
  Stander}, \bibinfo{person}{Eric Hallahan}, \bibinfo{person}{Louis
  Castricato}, {and} \bibinfo{person}{Edward Raff}.}
  \bibinfo{year}{2022}\natexlab{}.
\newblock \showarticletitle{VQGAN-CLIP: Open Domain Image Generation and
  Editing with Natural Language Guidance}.
\newblock \bibinfo{journal}{\emph{arXiv preprint arXiv:2204.08583}}
  (\bibinfo{year}{2022}).
\newblock


\bibitem[Darmetko(2021)]%
        {Darmetko2021fake}
\bibfield{author}{\bibinfo{person}{Tomasz Darmetko}.}
  \bibinfo{year}{2021}\natexlab{}.
\newblock \bibinfo{title}{Fake or not? {Generating} adversarial examples from
  language models}.
\newblock \bibinfo{howpublished}{Doctoral thesis submitted to Maastricht
  University}.
\newblock


\bibitem[Dawson et~al\mbox{.}(2020)]%
        {dawson2020can}
\bibfield{author}{\bibinfo{person}{Phillip Dawson}, \bibinfo{person}{Wendy
  Sutherland-Smith}, {and} \bibinfo{person}{Mark Ricksen}.}
  \bibinfo{year}{2020}\natexlab{}.
\newblock \showarticletitle{Can software improve marker accuracy at detecting
  contract cheating? A pilot study of the {Turnitin} authorship investigate
  alpha}.
\newblock \bibinfo{journal}{\emph{Assessment \& Evaluation in higher
  education}} \bibinfo{volume}{45}, \bibinfo{number}{4} (\bibinfo{year}{2020}),
  \bibinfo{pages}{473--482}.
\newblock


\bibitem[Devore-McDonald and Berger(2020)]%
        {devore2020mossad}
\bibfield{author}{\bibinfo{person}{Breanna Devore-McDonald} {and}
  \bibinfo{person}{Emery~D Berger}.} \bibinfo{year}{2020}\natexlab{}.
\newblock \showarticletitle{Mossad: defeating software plagiarism detection}.
\newblock \bibinfo{journal}{\emph{Proceedings of the ACM on Programming
  Languages}} \bibinfo{volume}{4}, \bibinfo{number}{OOPSLA}
  (\bibinfo{year}{2020}), \bibinfo{pages}{1--28}.
\newblock


\bibitem[Drori et~al\mbox{.}(2021)]%
        {drori2021neural}
\bibfield{author}{\bibinfo{person}{Iddo Drori}, \bibinfo{person}{Sunny Tran},
  \bibinfo{person}{Roman Wang}, \bibinfo{person}{Newman Cheng},
  \bibinfo{person}{Kevin Liu}, \bibinfo{person}{Leonard Tang},
  \bibinfo{person}{Elizabeth Ke}, \bibinfo{person}{Nikhil Singh},
  \bibinfo{person}{Taylor~L Patti}, \bibinfo{person}{Jayson Lynch},
  {et~al\mbox{.}}} \bibinfo{year}{2021}\natexlab{}.
\newblock \showarticletitle{A Neural Network Solves and Generates Mathematics
  Problems by Program Synthesis: Calculus, Differential Equations, Linear
  Algebra, and More}.
\newblock \bibinfo{journal}{\emph{arXiv preprint arXiv:2112.15594}}
  (\bibinfo{year}{2021}).
\newblock


\bibitem[Drori and Verma(2021)]%
        {drori2021solving}
\bibfield{author}{\bibinfo{person}{Iddo Drori} {and} \bibinfo{person}{Nakul
  Verma}.} \bibinfo{year}{2021}\natexlab{}.
\newblock \showarticletitle{Solving Linear Algebra by Program Synthesis}.
\newblock \bibinfo{journal}{\emph{arXiv preprint arXiv:2111.08171}}
  (\bibinfo{year}{2021}).
\newblock


\bibitem[Eret and Gokmenoglu(2010)]%
        {eret2010plagiarism}
\bibfield{author}{\bibinfo{person}{Esra Eret} {and} \bibinfo{person}{Tuba
  Gokmenoglu}.} \bibinfo{year}{2010}\natexlab{}.
\newblock \showarticletitle{Plagiarism in higher education: A case study with
  prospective academicians}.
\newblock \bibinfo{journal}{\emph{Procedia-Social and Behavioral Sciences}}
  \bibinfo{volume}{2}, \bibinfo{number}{2} (\bibinfo{year}{2010}),
  \bibinfo{pages}{3303--3307}.
\newblock


\bibitem[Faidhi and Robinson(1987)]%
        {faidhi1987empirical}
\bibfield{author}{\bibinfo{person}{Jinan~AW Faidhi} {and}
  \bibinfo{person}{Stuart~K Robinson}.} \bibinfo{year}{1987}\natexlab{}.
\newblock \showarticletitle{An empirical approach for detecting program
  similarity and plagiarism within a university programming environment}.
\newblock \bibinfo{journal}{\emph{Computers \& Education}}
  \bibinfo{volume}{11}, \bibinfo{number}{1} (\bibinfo{year}{1987}),
  \bibinfo{pages}{11--19}.
\newblock


\bibitem[Feldman(2020)]%
        {feldman2020does}
\bibfield{author}{\bibinfo{person}{Vitaly Feldman}.}
  \bibinfo{year}{2020}\natexlab{}.
\newblock \showarticletitle{Does learning require memorization? a short tale
  about a long tail}. In \bibinfo{booktitle}{\emph{Proceedings of the 52nd
  Annual ACM SIGACT Symposium on Theory of Computing}}.
  \bibinfo{pages}{954--959}.
\newblock


\bibitem[Folt{\`y}nek et~al\mbox{.}(2020)]%
        {foltynek2020detecting}
\bibfield{author}{\bibinfo{person}{Tom{\'a}{\v{s}} Folt{\`y}nek},
  \bibinfo{person}{Terry Ruas}, \bibinfo{person}{Philipp Scharpf},
  \bibinfo{person}{Norman Meuschke}, \bibinfo{person}{Moritz Schubotz},
  \bibinfo{person}{William Grosky}, {and} \bibinfo{person}{Bela Gipp}.}
  \bibinfo{year}{2020}\natexlab{}.
\newblock \showarticletitle{Detecting machine-obfuscated plagiarism}. In
  \bibinfo{booktitle}{\emph{International Conference on Information}}.
  Springer, \bibinfo{pages}{816--827}.
\newblock


\bibitem[Fried et~al\mbox{.}(2022)]%
        {fried2022incoder}
\bibfield{author}{\bibinfo{person}{Daniel Fried}, \bibinfo{person}{Armen
  Aghajanyan}, \bibinfo{person}{Jessy Lin}, \bibinfo{person}{Sida Wang},
  \bibinfo{person}{Eric Wallace}, \bibinfo{person}{Freda Shi},
  \bibinfo{person}{Ruiqi Zhong}, \bibinfo{person}{Wen-tau Yih},
  \bibinfo{person}{Luke Zettlemoyer}, {and} \bibinfo{person}{Mike Lewis}.}
  \bibinfo{year}{2022}\natexlab{}.
\newblock \showarticletitle{InCoder: A Generative Model for Code Infilling and
  Synthesis}.
\newblock \bibinfo{journal}{\emph{arXiv preprint arXiv:2204.05999}}
  (\bibinfo{year}{2022}).
\newblock


\bibitem[Gao et~al\mbox{.}(2020)]%
        {gao2020pile}
\bibfield{author}{\bibinfo{person}{Leo Gao}, \bibinfo{person}{Stella Biderman},
  \bibinfo{person}{Sid Black}, \bibinfo{person}{Laurence Golding},
  \bibinfo{person}{Travis Hoppe}, \bibinfo{person}{Charles Foster},
  \bibinfo{person}{Jason Phang}, \bibinfo{person}{Horace He},
  \bibinfo{person}{Anish Thite}, \bibinfo{person}{Noa Nabeshima},
  \bibinfo{person}{Shawn Presser}, {and} \bibinfo{person}{Connor Leahy}.}
  \bibinfo{year}{2020}\natexlab{}.
\newblock \showarticletitle{{The Pile: An 800GB Dataset of Diverse Text for
  Language Modeling}}.
\newblock \bibinfo{journal}{\emph{arXiv preprint arXiv:2101.00027}}
  (\bibinfo{year}{2020}).
\newblock


\bibitem[Han et~al\mbox{.}(2022)]%
        {han2022meet}
\bibfield{author}{\bibinfo{person}{Seungju Han}, \bibinfo{person}{Beomsu Kim},
  \bibinfo{person}{Jin~Yong Yoo}, \bibinfo{person}{Seokjun Seo},
  \bibinfo{person}{Sangbum Kim}, \bibinfo{person}{Enkhbayar Erdenee}, {and}
  \bibinfo{person}{Buru Chang}.} \bibinfo{year}{2022}\natexlab{}.
\newblock \showarticletitle{Meet Your Favorite Character: Open-domain Chatbot
  Mimicking Fictional Characters with only a Few Utterances}.
\newblock \bibinfo{journal}{\emph{arXiv preprint arXiv:2204.10825}}
  (\bibinfo{year}{2022}).
\newblock


\bibitem[He et~al\mbox{.}(2021)]%
        {he2021model}
\bibfield{author}{\bibinfo{person}{Xuanli He}, \bibinfo{person}{Lingjuan Lyu},
  \bibinfo{person}{Lichao Sun}, {and} \bibinfo{person}{Qiongkai Xu}.}
  \bibinfo{year}{2021}\natexlab{}.
\newblock \showarticletitle{Model Extraction and Adversarial Transferability,
  Your {BERT} is Vulnerable!}. In \bibinfo{booktitle}{\emph{Proceedings of the
  2021 Conference of the North American Chapter of the Association for
  Computational Linguistics: Human Language Technologies}}.
  \bibinfo{pages}{2006--2012}.
\newblock


\bibitem[Hendrycks et~al\mbox{.}(2021)]%
        {hendrycks2021measuring}
\bibfield{author}{\bibinfo{person}{Dan Hendrycks}, \bibinfo{person}{Steven
  Basart}, \bibinfo{person}{Saurav Kadavath}, \bibinfo{person}{Mantas Mazeika},
  \bibinfo{person}{Akul Arora}, \bibinfo{person}{Ethan Guo},
  \bibinfo{person}{Collin Burns}, \bibinfo{person}{Samir Puranik},
  \bibinfo{person}{Horace He}, \bibinfo{person}{Dawn Song}, {and}
  \bibinfo{person}{Jacob Steinhardt}.} \bibinfo{year}{2021}\natexlab{}.
\newblock \showarticletitle{Measuring Coding Challenge Competence With {APPS}}.
  In \bibinfo{booktitle}{\emph{Thirty-fifth Conference on Neural Information
  Processing Systems Datasets and Benchmarks Track}}.
\newblock


\bibitem[Holtzman et~al\mbox{.}(2019)]%
        {holtzman2019curious}
\bibfield{author}{\bibinfo{person}{Ari Holtzman}, \bibinfo{person}{Jan Buys},
  \bibinfo{person}{Li Du}, \bibinfo{person}{Maxwell Forbes}, {and}
  \bibinfo{person}{Yejin Choi}.} \bibinfo{year}{2019}\natexlab{}.
\newblock \showarticletitle{The curious case of neural text degeneration}.
\newblock \bibinfo{journal}{\emph{arXiv preprint arXiv:1904.09751}}
  (\bibinfo{year}{2019}).
\newblock


\bibitem[Hu et~al\mbox{.}(2020)]%
        {hu2020text}
\bibfield{author}{\bibinfo{person}{Zhiqiang Hu}, \bibinfo{person}{Roy Ka-Wei
  Lee}, \bibinfo{person}{Charu~C Aggarwal}, {and} \bibinfo{person}{Aston
  Zhang}.} \bibinfo{year}{2020}\natexlab{}.
\newblock \showarticletitle{Text Style Transfer: A Review and Experimental
  Evaluation}.
\newblock \bibinfo{journal}{\emph{arXiv preprint arXiv:2010.12742}}
  (\bibinfo{year}{2020}).
\newblock


\bibitem[Johnson(2022)]%
        {johnsonalgorithms}
\bibfield{author}{\bibinfo{person}{Gabbrielle Johnson}.}
  \bibinfo{year}{2022}\natexlab{}.
\newblock \showarticletitle{Are Algorithms Value-Free? Feminist Theoretical
  Virtues in Machine Learning}.
\newblock \bibinfo{journal}{\emph{Journal of Moral Philosophy, special issue on
  "Justice, Power, and the Ethics of Algorithmic Decision-Making"}}
  (\bibinfo{year}{2022}).
\newblock


\bibitem[Juliana et~al\mbox{.}(2020)]%
        {juliana2020turnitin}
\bibfield{author}{\bibinfo{person}{Juliana}, \bibinfo{person}{Rudy Pramono},
  \bibinfo{person}{Rizaldi Parani}, {and} \bibinfo{person}{Arifin
  Djakasaputra}.} \bibinfo{year}{2020}\natexlab{}.
\newblock \showarticletitle{Transformation Of Learning For Early Childhood
  Education Through Parent Assistance In The Pandemic Covid-19}.
\newblock \bibinfo{journal}{\emph{Transformation Of Learning For Early
  Childhood Education Through Parent Assistance In The Pandemic Covid-19}}
  \bibinfo{volume}{20}, \bibinfo{number}{XX} (\bibinfo{year}{2020}),
  \bibinfo{pages}{1--14}.
\newblock


\bibitem[Karnalim et~al\mbox{.}(2019)]%
        {karnalim2019source}
\bibfield{author}{\bibinfo{person}{Oscar Karnalim}, \bibinfo{person}{Setia
  Budi}, \bibinfo{person}{Hapnes Toba}, {and} \bibinfo{person}{Mike Joy}.}
  \bibinfo{year}{2019}\natexlab{}.
\newblock \showarticletitle{Source Code Plagiarism Detection in Academia with
  Information Retrieval: Dataset and the Observatjion.}
\newblock \bibinfo{journal}{\emph{Informatics in Education}}
  \bibinfo{volume}{18}, \bibinfo{number}{2} (\bibinfo{year}{2019}),
  \bibinfo{pages}{321--344}.
\newblock


\bibitem[Karnalim et~al\mbox{.}(2020)]%
        {karnalim2020choosing}
\bibfield{author}{\bibinfo{person}{Oscar Karnalim}, \bibinfo{person}{Judy
  Sheard}, \bibinfo{person}{Ilir Dema}, \bibinfo{person}{Amey Karkare},
  \bibinfo{person}{Juho Leinonen}, \bibinfo{person}{Michael Liut}, {and}
  \bibinfo{person}{Ren{\'e}e McCauley}.} \bibinfo{year}{2020}\natexlab{}.
\newblock \showarticletitle{Choosing code segments to exclude from code
  similarity detection}.
\newblock In \bibinfo{booktitle}{\emph{Proceedings of the Working Group Reports
  on Innovation and Technology in Computer Science Education}}.
  \bibinfo{pages}{1--19}.
\newblock


\bibitem[Krishna et~al\mbox{.}(2020)]%
        {krishna2020thieves}
\bibfield{author}{\bibinfo{person}{Kalpesh Krishna},
  \bibinfo{person}{Gaurav~Singh Tomar}, \bibinfo{person}{Ankur~P. Parikh},
  \bibinfo{person}{Nicolas Papernot}, {and} \bibinfo{person}{Mohit Iyyer}.}
  \bibinfo{year}{2020}\natexlab{}.
\newblock \showarticletitle{{Thieves on Sesame Street! Model Extraction of
  BERT-based APIs}}. In \bibinfo{booktitle}{\emph{International Conference on
  Learning Representations}}.
\newblock


\bibitem[Lancaster and Culwin(2004)]%
        {lancaster2004comparison}
\bibfield{author}{\bibinfo{person}{Thomas Lancaster} {and}
  \bibinfo{person}{Fintan Culwin}.} \bibinfo{year}{2004}\natexlab{}.
\newblock \showarticletitle{A comparison of source code plagiarism detection
  engines}.
\newblock \bibinfo{journal}{\emph{Computer Science Education}}
  \bibinfo{volume}{14}, \bibinfo{number}{2} (\bibinfo{year}{2004}),
  \bibinfo{pages}{101--112}.
\newblock


\bibitem[Leahy and Biderman(2021)]%
        {leahy2021hard}
\bibfield{author}{\bibinfo{person}{Connor Leahy} {and} \bibinfo{person}{Stella
  Biderman}.} \bibinfo{year}{2021}\natexlab{}.
\newblock \showarticletitle{The Hard Problem of Aligning {AI} to Human Values}.
\newblock In \bibinfo{booktitle}{\emph{The State of AI Ethics Report (Volume
  4)}}.
\newblock


\bibitem[Lee et~al\mbox{.}(2022b)]%
        {lee2022coauthor}
\bibfield{author}{\bibinfo{person}{Mina Lee}, \bibinfo{person}{Percy Liang},
  {and} \bibinfo{person}{Qian Yang}.} \bibinfo{year}{2022}\natexlab{b}.
\newblock \showarticletitle{CoAuthor: Designing a Human-AI Collaborative
  Writing Dataset for Exploring Language Model Capabilities}.
\newblock \bibinfo{journal}{\emph{arXiv preprint arXiv:2201.06796}}
  (\bibinfo{year}{2022}).
\newblock


\bibitem[Lee et~al\mbox{.}(2022a)]%
        {lee2022interactive}
\bibfield{author}{\bibinfo{person}{Yoonjoo Lee}, \bibinfo{person}{Tae~Soo Kim},
  \bibinfo{person}{Minsuk Chang}, {and} \bibinfo{person}{Juho Kim}.}
  \bibinfo{year}{2022}\natexlab{a}.
\newblock \showarticletitle{Interactive Children’s Story Rewriting Through
  Parent-Children Interaction}. In \bibinfo{booktitle}{\emph{Proceedings of the
  First Workshop on Intelligent and Interactive Writing Assistants (In2Writing
  2022)}}. \bibinfo{pages}{62--71}.
\newblock


\bibitem[Li et~al\mbox{.}(2020)]%
        {li2020leveraging}
\bibfield{author}{\bibinfo{person}{Wei Li}, \bibinfo{person}{Xinyan Xiao},
  \bibinfo{person}{Jiachen Liu}, \bibinfo{person}{Hua Wu},
  \bibinfo{person}{Haifeng Wang}, {and} \bibinfo{person}{Junping Du}.}
  \bibinfo{year}{2020}\natexlab{}.
\newblock \showarticletitle{Leveraging Graph to Improve Abstractive
  Multi-Document Summarization}. In \bibinfo{booktitle}{\emph{Proceedings of
  the 58th Annual Meeting of the Association for Computational Linguistics}}.
  \bibinfo{pages}{6232--6243}.
\newblock


\bibitem[Li and Liang(2021)]%
        {li2021prefix}
\bibfield{author}{\bibinfo{person}{Xiang~Lisa Li} {and} \bibinfo{person}{Percy
  Liang}.} \bibinfo{year}{2021}\natexlab{}.
\newblock \showarticletitle{Prefix-tuning: Optimizing continuous prompts for
  generation}.
\newblock \bibinfo{journal}{\emph{arXiv preprint arXiv:2101.00190}}
  (\bibinfo{year}{2021}).
\newblock


\bibitem[Liu et~al\mbox{.}(2021)]%
        {liu2021pre}
\bibfield{author}{\bibinfo{person}{Pengfei Liu}, \bibinfo{person}{Weizhe Yuan},
  \bibinfo{person}{Jinlan Fu}, \bibinfo{person}{Zhengbao Jiang},
  \bibinfo{person}{Hiroaki Hayashi}, {and} \bibinfo{person}{Graham Neubig}.}
  \bibinfo{year}{2021}\natexlab{}.
\newblock \showarticletitle{Pre-train, prompt, and predict: A systematic survey
  of prompting methods in natural language processing}.
\newblock \bibinfo{journal}{\emph{arXiv preprint arXiv:2107.13586}}
  (\bibinfo{year}{2021}).
\newblock


\bibitem[Liu and Lapata(2019)]%
        {liu2019text}
\bibfield{author}{\bibinfo{person}{Yang Liu} {and} \bibinfo{person}{Mirella
  Lapata}.} \bibinfo{year}{2019}\natexlab{}.
\newblock \showarticletitle{Text Summarization with Pretrained Encoders}. In
  \bibinfo{booktitle}{\emph{Proceedings of the 2019 Conference on Empirical
  Methods in Natural Language Processing and the 9th International Joint
  Conference on Natural Language Processing (EMNLP-IJCNLP)}}.
  \bibinfo{pages}{3730--3740}.
\newblock


\bibitem[Luke et~al\mbox{.}(2014)]%
        {luke2014software}
\bibfield{author}{\bibinfo{person}{Divya Luke}, \bibinfo{person}{PS Divya},
  \bibinfo{person}{Sony~L Johnson}, \bibinfo{person}{S Sreeprabha}, {and}
  \bibinfo{person}{Elizabeth~B Varghese}.} \bibinfo{year}{2014}\natexlab{}.
\newblock \showarticletitle{Software plagiarism detection techniques: A
  comparative study}. In \bibinfo{booktitle}{\emph{2018 5th International
  Symposium on Emerging Trends and Technologies in Libraries and Information
  Services (ETTLIS)}}. \bibinfo{publisher}{Citeseer}.
\newblock


\bibitem[Mukherjee et~al\mbox{.}(2021)]%
        {mukherjee2021neural}
\bibfield{author}{\bibinfo{person}{Rohan Mukherjee}, \bibinfo{person}{Yeming
  Wen}, \bibinfo{person}{Dipak Chaudhari}, \bibinfo{person}{Thomas Reps},
  \bibinfo{person}{Swarat Chaudhuri}, {and} \bibinfo{person}{Chris Jermaine}.}
  \bibinfo{year}{2021}\natexlab{}.
\newblock \showarticletitle{Neural Program Generation Modulo Static Analysis}.
  In \bibinfo{booktitle}{\emph{Thirty-Fifth Conference on Neural Information
  Processing Systems}}.
\newblock


\bibitem[Narayan et~al\mbox{.}(2018)]%
        {narayan2018ranking}
\bibfield{author}{\bibinfo{person}{Shashi Narayan}, \bibinfo{person}{Shay~B
  Cohen}, {and} \bibinfo{person}{Mirella Lapata}.}
  \bibinfo{year}{2018}\natexlab{}.
\newblock \showarticletitle{Ranking Sentences for Extractive Summarization with
  Reinforcement Learning}. In \bibinfo{booktitle}{\emph{Proceedings of the 2018
  Conference of the North American Chapter of the Association for Computational
  Linguistics: Human Language Technologies, Volume 1 (Long Papers)}}.
  \bibinfo{pages}{1747--1759}.
\newblock


\bibitem[Oppenlaender(2022)]%
        {oppenlaender2022prompt}
\bibfield{author}{\bibinfo{person}{Jonas Oppenlaender}.}
  \bibinfo{year}{2022}\natexlab{}.
\newblock \showarticletitle{Prompt Engineering for Text-Based Generative Art}.
\newblock \bibinfo{journal}{\emph{arXiv preprint arXiv:2204.13988}}
  (\bibinfo{year}{2022}).
\newblock


\bibitem[Pawelczak(2018)]%
        {pawelczak2018benefits}
\bibfield{author}{\bibinfo{person}{Dieter Pawelczak}.}
  \bibinfo{year}{2018}\natexlab{}.
\newblock \showarticletitle{Benefits and drawbacks of source code plagiarism
  detection in engineering education}. In \bibinfo{booktitle}{\emph{2018 IEEE
  Global Engineering Education Conference (EDUCON)}}. IEEE,
  \bibinfo{pages}{1048--1056}.
\newblock


\bibitem[Raffel et~al\mbox{.}(2019)]%
        {raffel2019exploring}
\bibfield{author}{\bibinfo{person}{Colin Raffel}, \bibinfo{person}{Noam
  Shazeer}, \bibinfo{person}{Adam Roberts}, \bibinfo{person}{Katherine Lee},
  \bibinfo{person}{Sharan Narang}, \bibinfo{person}{Michael Matena},
  \bibinfo{person}{Yanqi Zhou}, \bibinfo{person}{Wei Li}, {and}
  \bibinfo{person}{Peter~J Liu}.} \bibinfo{year}{2019}\natexlab{}.
\newblock \showarticletitle{Exploring the limits of transfer learning with a
  unified text-to-text transformer}.
\newblock \bibinfo{journal}{\emph{arXiv preprint arXiv:1910.10683}}
  (\bibinfo{year}{2019}).
\newblock


\bibitem[Reynolds and McDonell(2021)]%
        {reynolds2021prompt}
\bibfield{author}{\bibinfo{person}{Laria Reynolds} {and} \bibinfo{person}{Kyle
  McDonell}.} \bibinfo{year}{2021}\natexlab{}.
\newblock \showarticletitle{Prompt programming for large language models:
  Beyond the few-shot paradigm}. In \bibinfo{booktitle}{\emph{Extended
  Abstracts of the 2021 CHI Conference on Human Factors in Computing Systems}}.
  \bibinfo{pages}{1--7}.
\newblock


\bibitem[Sanh et~al\mbox{.}(2022)]%
        {sanh2022multitask}
\bibfield{author}{\bibinfo{person}{Victor Sanh}, \bibinfo{person}{Albert
  Webson}, \bibinfo{person}{Colin Raffel}, \bibinfo{person}{Stephen Bach},
  \bibinfo{person}{Lintang Sutawika}, \bibinfo{person}{Zaid Alyafeai},
  \bibinfo{person}{Antoine Chaffin}, \bibinfo{person}{Arnaud Stiegler},
  \bibinfo{person}{Teven Le~Scao}, \bibinfo{person}{Arun Raja},
  {et~al\mbox{.}}} \bibinfo{year}{2022}\natexlab{}.
\newblock \showarticletitle{Multitask Prompted Training Enables Zero-Shot Task
  Generalization}. In \bibinfo{booktitle}{\emph{The Tenth International
  Conference on Learning Representations}}.
\newblock


\bibitem[Sheahen and Joyner(2016)]%
        {sheahen2016taps}
\bibfield{author}{\bibinfo{person}{Dana Sheahen} {and} \bibinfo{person}{David
  Joyner}.} \bibinfo{year}{2016}\natexlab{}.
\newblock \showarticletitle{{TAPS:} A {MOSS} extension for detecting software
  plagiarism at scale}. In \bibinfo{booktitle}{\emph{Proceedings of the Third
  (2016) ACM Conference on Learning@ Scale}}. \bibinfo{pages}{285--288}.
\newblock


\bibitem[Sheikh(2022)]%
        {sheikhteaching}
\bibfield{author}{\bibinfo{person}{Waseem Sheikh}.}
  \bibinfo{year}{2022}\natexlab{}.
\newblock \showarticletitle{Teaching C++ programming using automated unit
  testing and test-driven development—Design and efficacy study}.
\newblock \bibinfo{journal}{\emph{Computer Applications in Engineering
  Education}} (\bibinfo{year}{2022}).
\newblock


\bibitem[Shporer et~al\mbox{.}(2021)]%
        {shporerlearning}
\bibfield{author}{\bibinfo{person}{Avi Shporer}, \bibinfo{person}{Sunny Tran},
  \bibinfo{person}{Nikhil Singh}, \bibinfo{person}{Brandon Kates},
  \bibinfo{person}{Jayson Lynch}, {and} \bibinfo{person}{Iddo Drori}.}
  \bibinfo{year}{2021}\natexlab{}.
\newblock \showarticletitle{Learning Methods for Solving Astronomy Course
  Problems}.
\newblock \bibinfo{journal}{\emph{preprint}} (\bibinfo{year}{2021}).
\newblock
\urldef\tempurl%
\url{https://www.cs.columbia.edu/~idrori/Learning_Methods_for_Solving_Astronomy_Course_Problems.pdf}
\showURL{%
\tempurl}


\bibitem[Singh et~al\mbox{.}(2017)]%
        {singh2017gradescope}
\bibfield{author}{\bibinfo{person}{Arjun Singh}, \bibinfo{person}{Sergey
  Karayev}, \bibinfo{person}{Kevin Gutowski}, {and} \bibinfo{person}{Pieter
  Abbeel}.} \bibinfo{year}{2017}\natexlab{}.
\newblock \showarticletitle{Gradescope: a fast, flexible, and fair system for
  scalable assessment of handwritten work}. In
  \bibinfo{booktitle}{\emph{Proceedings of the fourth (2017) acm conference on
  learning@ scale}}. \bibinfo{pages}{81--88}.
\newblock


\bibitem[Tran et~al\mbox{.}(2021)]%
        {tran2021solving}
\bibfield{author}{\bibinfo{person}{Sunny Tran}, \bibinfo{person}{Pranav
  Krishna}, \bibinfo{person}{Ishan Pakuwal}, \bibinfo{person}{Prabhakar Kafle},
  \bibinfo{person}{Nikhil Singh}, \bibinfo{person}{Jayson Lynch}, {and}
  \bibinfo{person}{Iddo Drori}.} \bibinfo{year}{2021}\natexlab{}.
\newblock \showarticletitle{Solving machine learning problems}. In
  \bibinfo{booktitle}{\emph{Asian Conference on Machine Learning}}. PMLR,
  \bibinfo{pages}{470--485}.
\newblock


\bibitem[Underwood(2022)]%
        {underwood2021mapping}
\bibfield{author}{\bibinfo{person}{Ted Underwood}.}
  \bibinfo{year}{2022}\natexlab{}.
\newblock \showarticletitle{Mapping the Latent Spaces of Culture}.
\newblock \bibinfo{journal}{\emph{To appear in Startwords}}
  (\bibinfo{year}{2022}).
\newblock
\urldef\tempurl%
\url{https://doi.org/10.17613/faaa-1r21}
\showDOI{\tempurl}


\bibitem[Vaithilingam et~al\mbox{.}(2022)]%
        {vaithilingam2022expectation}
\bibfield{author}{\bibinfo{person}{Priyan Vaithilingam},
  \bibinfo{person}{Tianyi Zhang}, {and} \bibinfo{person}{Elena~L Glassman}.}
  \bibinfo{year}{2022}\natexlab{}.
\newblock \showarticletitle{Expectation vs. Experience: Evaluating the
  Usability of Code Generation Tools Powered by Large Language Models}. In
  \bibinfo{booktitle}{\emph{CHI Conference on Human Factors in Computing
  Systems Extended Abstracts}}. \bibinfo{pages}{1--7}.
\newblock


\bibitem[Wang and Komatsuzaki(2021)]%
        {gpt-j}
\bibfield{author}{\bibinfo{person}{Ben Wang} {and} \bibinfo{person}{Aran
  Komatsuzaki}.} \bibinfo{year}{2021}\natexlab{}.
\newblock \bibinfo{title}{{GPT-J-6B:} A 6 Billion Parameter Autoregressive
  Language Model}.
\newblock
\newblock


\bibitem[Weisz et~al\mbox{.}(2022)]%
        {weisz2022better}
\bibfield{author}{\bibinfo{person}{Justin~D Weisz}, \bibinfo{person}{Michael
  Muller}, \bibinfo{person}{Steven~I Ross}, \bibinfo{person}{Fernando
  Martinez}, \bibinfo{person}{Stephanie Houde}, \bibinfo{person}{Mayank
  Agarwal}, \bibinfo{person}{Kartik Talamadupula}, {and}
  \bibinfo{person}{John~T Richards}.} \bibinfo{year}{2022}\natexlab{}.
\newblock \showarticletitle{Better together? an evaluation of ai-supported code
  translation}. In \bibinfo{booktitle}{\emph{27th International Conference on
  Intelligent User Interfaces}}. \bibinfo{pages}{369--391}.
\newblock


\bibitem[Wu et~al\mbox{.}(2021)]%
        {wu2021ai}
\bibfield{author}{\bibinfo{person}{Zhuohao Wu}, \bibinfo{person}{Danwen Ji},
  \bibinfo{person}{Kaiwen Yu}, \bibinfo{person}{Xianxu Zeng},
  \bibinfo{person}{Dingming Wu}, {and} \bibinfo{person}{Mohammad Shidujaman}.}
  \bibinfo{year}{2021}\natexlab{}.
\newblock \showarticletitle{AI Creativity and the Human-AI Co-creation Model}.
  In \bibinfo{booktitle}{\emph{International Conference on Human-Computer
  Interaction}}. Springer, \bibinfo{pages}{171--190}.
\newblock


\bibitem[Xu et~al\mbox{.}(2022)]%
        {xu2022systematic}
\bibfield{author}{\bibinfo{person}{Frank~F Xu}, \bibinfo{person}{Uri Alon},
  \bibinfo{person}{Graham Neubig}, {and} \bibinfo{person}{Vincent~J
  Hellendoorn}.} \bibinfo{year}{2022}\natexlab{}.
\newblock \showarticletitle{A systematic evaluation of large language models of
  code}.
\newblock \bibinfo{journal}{\emph{arXiv preprint arXiv:2202.13169}}
  (\bibinfo{year}{2022}).
\newblock


\bibitem[Yasid et~al\mbox{.}(2020)]%
        {yasid2020plagiarism}
\bibfield{author}{\bibinfo{person}{Muhammad Yasid},
  \bibinfo{person}{Gomgom~T.P. Siregar}, {and} \bibinfo{person}{Muhammad~Ansori
  Lubis}.} \bibinfo{year}{2020}\natexlab{}.
\newblock \showarticletitle{The Policy of Credit Payment Relaxation in
  Overcoming the Impact of {Covid-19} Spread to the Economic Society in
  {Indonesia}}.
\newblock \bibinfo{journal}{\emph{Journal of Advanced Research in Dynamical and
  Control Systems}} (\bibinfo{year}{2020}).
\newblock


\bibitem[Yuan et~al\mbox{.}(2022)]%
        {yuan2022wordcraft}
\bibfield{author}{\bibinfo{person}{Ann Yuan}, \bibinfo{person}{Andy Coenen},
  \bibinfo{person}{Emily Reif}, {and} \bibinfo{person}{Daphne Ippolito}.}
  \bibinfo{year}{2022}\natexlab{}.
\newblock \showarticletitle{Wordcraft: Story Writing With Large Language
  Models}. In \bibinfo{booktitle}{\emph{27th International Conference on
  Intelligent User Interfaces}}. \bibinfo{pages}{841--852}.
\newblock


\bibitem[Zhang et~al\mbox{.}(2022)]%
        {zhang2022opt}
\bibfield{author}{\bibinfo{person}{Susan Zhang}, \bibinfo{person}{Stephen
  Roller}, \bibinfo{person}{Naman Goyal}, \bibinfo{person}{Mikel Artetxe},
  \bibinfo{person}{Moya Chen}, \bibinfo{person}{Shuohui Chen},
  \bibinfo{person}{Christopher Dewan}, \bibinfo{person}{Mona Diab},
  \bibinfo{person}{Xian Li}, \bibinfo{person}{Xi~Victoria Lin},
  {et~al\mbox{.}}} \bibinfo{year}{2022}\natexlab{}.
\newblock \showarticletitle{OPT: Open Pre-trained Transformer Language Models}.
\newblock \bibinfo{journal}{\emph{arXiv preprint arXiv:2205.01068}}
  (\bibinfo{year}{2022}).
\newblock


\end{thebibliography}
\end{document}